\documentclass[Afour,sageh,times]{sagej}

\usepackage{moreverb,url}
\usepackage{pifont} 

\usepackage[hidelinks,colorlinks,bookmarksopen,bookmarksnumbered,citecolor=blue,urlcolor=blue,allcolors=blue,pdfborder={0 0 0}]{hyperref}

\hypersetup{
    colorlinks=true, 
    linkcolor=blue, 
    citecolor=cyan, 
    filecolor=blue, 
    urlcolor=blue,   
    pdfborder={0 0 0} 
}
\usepackage{xcolor}
\colorlet{linkequation}{blue}
\usepackage[colorlinks]{hyperref}
\usepackage{cleveref}

\usepackage{todonotes}
\usepackage{enumerate}
\usepackage{enumitem}
\usepackage[super]{nth}
\usepackage[amssymb]{SIunits}
\usepackage{graphicx}
\usepackage{nicefrac}
\usepackage{amssymb}
\usepackage{amsthm}
\usepackage{bm} 
\usepackage{footnote}
\usepackage{threeparttable}
\usepackage{epsfig}
\usepackage{graphicx}
\usepackage{balance}
\usepackage{booktabs}
\usepackage{subcaption}
\usepackage{wrapfig}
\usepackage{caption}
\usepackage{ulem}
\usepackage{ragged2e}
\usepackage{multirow}
\usepackage{array}
\usepackage{soul}
\usepackage{stfloats}
\usepackage{makecell}
\usepackage{algorithm}
\usepackage{algpseudocode}

\usepackage{todonotes}
\presetkeys{todonotes}{size=\tiny}{}

\newcommand{\fref}[1]{Figure~\ref{#1}}
\newcommand{\sref}[1]{Section~\ref{#1}}
\newcommand{\tref}[1]{Table~\ref{#1}}
\newcommand{\aref}[1]{Appendix~\ref{#1}}
\newcommand{\algorithmref}[1]{Algorithm~\ref{#1}}

\newcommand{\myparagraph}[1]{\noindent\textbf{#1}}

\let\oldequation\equation
\let\oldendequation\endequation

\renewenvironment{equation}
  {\small\oldequation}
  {\oldendequation}

\let\oldalign\align
\let\oldendalign\endalign

\renewenvironment{align}
  {\small\oldalign}
  {\oldendalign}

\begin{document}

\runninghead{Fu et al.}

\title{\LARGE AnyNav: Visual Neuro-Symbolic Friction Learning for Off-road Navigation}

\author{Taimeng Fu\affilnum{1}, Zitong Zhan\affilnum{1}, Zhipeng Zhao\affilnum{1}, Yi Du\affilnum{1}, Shaoshu Su\affilnum{1}, Xiao Lin\affilnum{1}, Ehsan Tarkesh Esfahani\affilnum{1}, Karthik Dantu\affilnum{1}, Souma Chowdhury\affilnum{1}, and Chen Wang\affilnum{1}}

\affiliation{%
\affilnum{1}University at Buffalo, NY 14260, USA
}

\corrauth{Chen Wang, Spatial AI \& Robotics (SAIR) Lab, Department of Computer Science and Engineering, University at Buffalo, NY 14260, USA.}

\begin{abstract}

Off-road navigation is critical for a wide range of field robotics applications from planetary exploration to disaster response. 
However, it remains a longstanding challenge due to unstructured environments and the inherently complex terrain-vehicle interactions. 
Traditional physics-based methods struggle to accurately capture the nonlinear dynamics underlying these interactions, while purely data-driven approaches often overfit to specific motion patterns, vehicle geometries, or platforms, limiting their generalization in diverse, real-world scenarios.
To address these limitations, we introduce AnyNav, a vision-based friction estimation and navigation framework grounded in neuro-symbolic principles.
Our approach integrates neural networks for visual perception with symbolic physical models for reasoning about terrain-vehicle dynamics.
To enable learning in a self-supervised manner for real-world settings, we adopt the imperative learning paradigm, where a bilevel optimization framework leverages physics-based optimization for training the friction network. This explicit incorporation of physical reasoning substantially improves generalization across terrains, vehicle types, and operating conditions.
Leveraging the predicted friction coefficients, we further develop a physics-informed navigation system capable of generating physically feasible, time-efficient paths together with corresponding speed profiles.
We demonstrate that AnyNav seamlessly transfers from simulation to real-world robotic platforms, exhibiting strong robustness across different four-wheeled vehicles and diverse off-road environments.

\end{abstract}


\maketitle

\section{Introduction}

Off-road navigation is crucial for vehicles to effectively operate in unstructured and unpredictable environments without traditional road networks \citep{di2021survey}. It is vital for various robotic applications such as planetary exploration \citep{wang2024risk}, disaster response \citep{schwarz2017nimbro}, and agricultural automation \citep{zhang2024development}.

Unlike structured on-road environments, off-road terrains exhibit diverse and irregular features such as loose gravel, mud, sand, and vegetation, each affecting vehicle dynamics differently \citep{islam2022off}. These terrains introduce complicated nonlinear interactions between the vehicle's wheels and the terrain. Despite numerous efforts, accurately predicting vehicle-terrain interactions while ensuring generalizability across different terrain types and vehicle configurations remains an open challenge.

\begin{figure}[t]
    \centering
    \includegraphics[width=\linewidth]{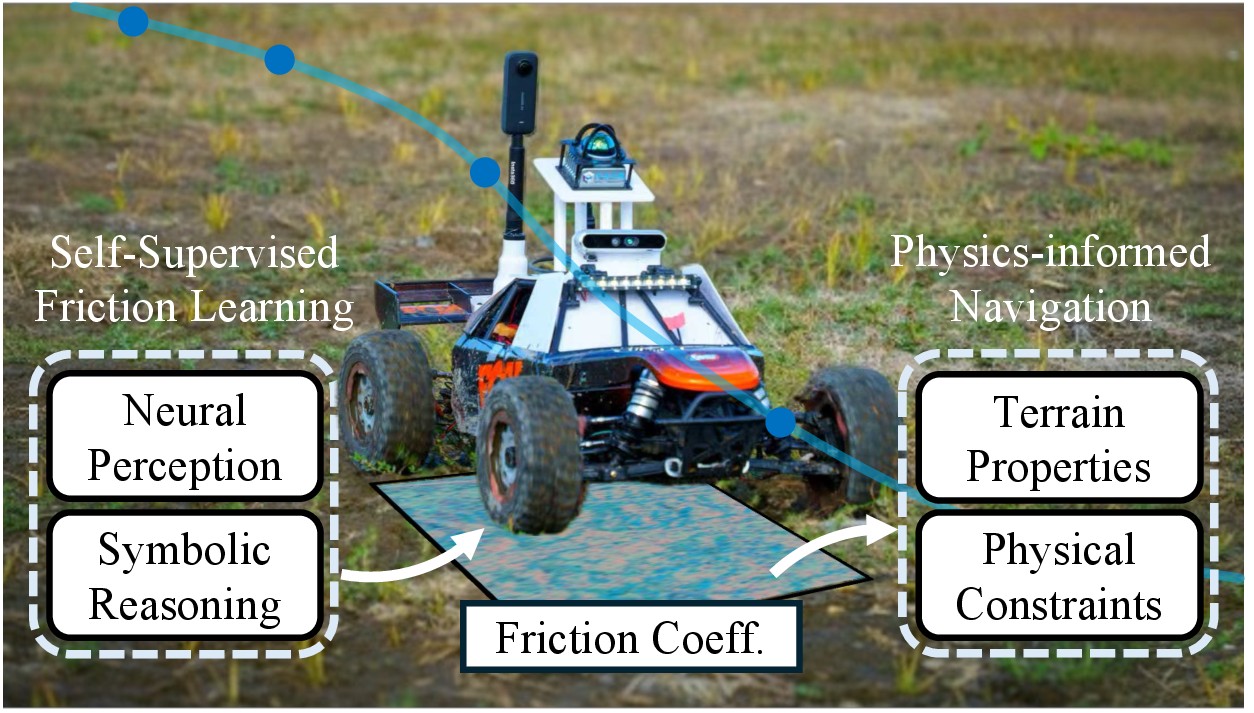}
    \caption{We tackle the challenges of off-road navigation by developing AnyNav, a neuro-symbolic framework for predicting friction coefficient and integrating these predictions as physical constraints in path and speed planning for reliable navigation.}
    \label{fig:head}
\end{figure}

Physics-based methods \citep{tanelli2012combined, zhu2016integrated} were developed to explicitly model the fundamental laws governing vehicle-terrain interactions. 
However, they cannot capture the complexities and variability of natural environments.
Learning-based techniques have also shown promise in end-to-end vehicle control by fitting the dynamics model from large datasets \citep{wang2023sa, wang2024pay}.
However, they often overfit to specific motion patterns and may produce unreasonable predictions when exposed to out-of-distribution scenarios, thereby limiting their generalizability.
Recent advancements in neuro-symbolic learning have infused physical laws into the learning process, enhancing system interpretability \citep{zhao2024physord, pokhrel2024cahsor}. Nevertheless, the learned dynamics models remain prone to overfitting to individual vehicles, as their neural components entangle terrain properties with robot configurations such as size, weight, and drive type.

Achieving generalizable and interpretable off-road navigation requires a comprehensive understanding of vehicle dynamics. A crucial factor is the friction coefficient, which plays a vital role in determining how a vehicle will interact with various terrain surfaces. 
However, obtaining ground-truth friction coefficients for training in real-world environments is notoriously difficult, which fundamentally limits the practicality and scalability of existing methods.
To address this challenge, one line of work classifies terrain into several predefined categories and assigns discrete friction coefficients to each class \citep{brandao2016material}; however, real-world terrain can exhibit out-of-distribution appearance and continuous friction values that deviate from the predefined categories, leading to limited generalization.
Another line of work employs tactile sensors mounted on robot arms to physically probe the ground and measure friction forces \citep{ewen2024you, le2021probabilistic}; however, these approaches provide only sparse samples and are infeasible for high-speed off-road driving or platforms that lack manipulators.

To address the lack of ground-truth friction labels, we introduce AnyNav, a self-supervised friction learning framework for off-road navigation. AnyNav is first trained in simulation and then transferred to the real world through a bilevel optimization scheme inspired by imperative learning \citep{wang2025imperative}. The framework combines (1) a neural perception network that infers terrain friction from visual input and (2) a symbolic reasoning engine grounded in physical laws that predicts vehicle motion from the estimated friction.
Self-supervised sim-to-real transfer is realized by optimizing the perception network at the upper level while enforcing consistency with physics-based optimization for motion predictions at the lower level. 
Enforcing consistency between predicted friction and the resulting measured vehicle motion enables data-efficient learning without annotated labels.

Additionally, to achieve reliable off-road navigation, we introduce a physics-informed navigation system that leverages the learned friction to enhance terrain analysis. By jointly considering friction properties, terrain geometry, and vehicle capabilities through physics-based costs and constraints, the planner generates physically feasible trajectories and avoids implausible scenarios such as climbing a hill with insufficient friction or driving at high speeds over rough terrain.
\fref{fig:head} provides an overview of how AnyNav connects visual friction prediction with path and speed planning.
Our contributions include:
\begin{itemize}[noitemsep,topsep=0pt,leftmargin=20pt]
    \item We propose a neuro-symbolic framework for visual friction learning, offering greater interpretability and improved generalizability in complex and unstructured environments.
    This framework, referred to as AnyNav, grounds a visual neural network in physical reasoning to predict friction coefficients. Its self-supervised design based on bilevel optimization enables adaptability and generalizability across diverse off-road terrains.
    \item We develop a terrain property map together with a physics-informed navigation system that generates paths and speed profiles under constraints of terrain friction, elevation, roughness, route distance, and vehicle speed.
    Extensive experiments in both simulation and real-world settings demonstrate the robustness and reliability of our navigation system across various off-road scenarios and vehicle platforms.
\end{itemize}

\section{Related Works}

\subsection{Physics-based Methods}

Physics-based approaches have been widely used for modeling vehicle dynamics in both structured and unstructured environments, by explicitly incorporating fundamental physical principles \citep{di2021survey}. The kinematic bicycle model \citep{ailon2005controllability, polack2018guaranteeing} leverages Newtonian mechanics to predict vehicle state transitions effectively in controlled scenarios. Enhancements like the Kalman Filter \citep{reina2019vehicle} and Monte Carlo simulations \citep{wu2015new} introduce stochastic elements to manage uncertainties, particularly in off-road conditions. However, these models often rely on idealized assumptions and simplified dynamics, which limit their applicability in complex, highly non-linear terrains.

A parallel line of research focuses on modeling tire-terrain interactions \citep{taheri2015technical}. These methods often employ principles of soil mechanics and account for tire deformation, as demonstrated by \cite{madsen2012physics, wasfy2018prediction, serban2017co}. While effective under certain conditions, these models typically depend on predefined parameters, reducing their adaptability to dynamic and heterogeneous environments. \cite{tanelli2012combined} integrated empirical observations with theoretical frameworks to estimate certain parameters, but their approach was limited to straight-line driving.

\begin{figure*}[t]
    \centering
    \includegraphics[width=\linewidth]{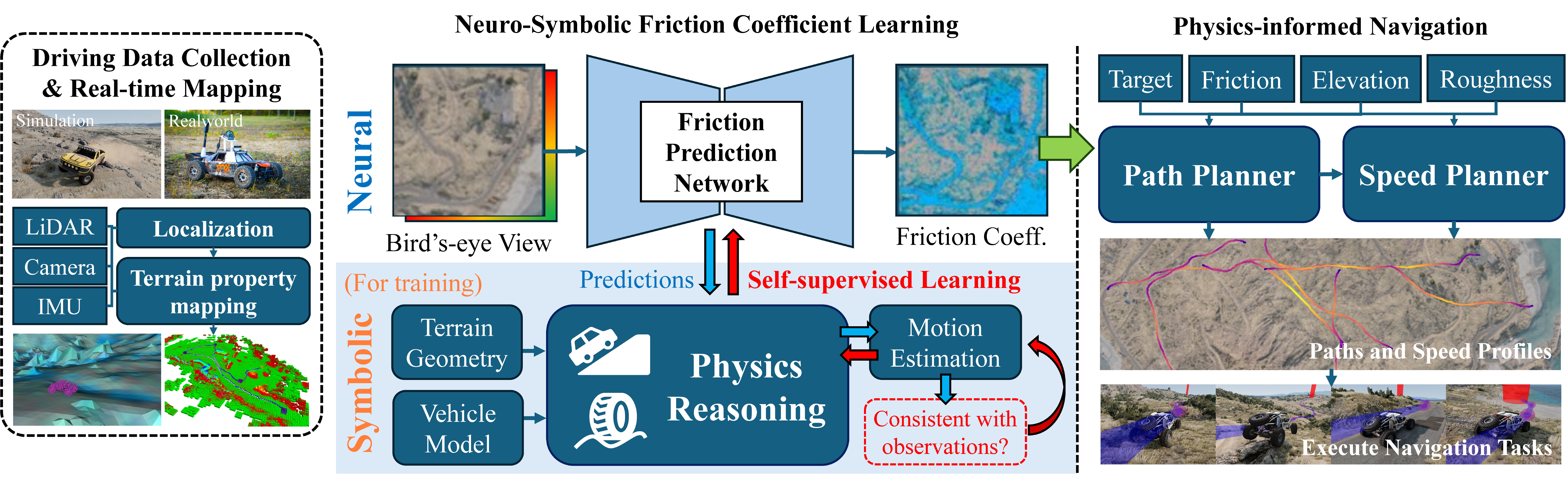}
    \caption{\textbf{AnyNav comprises a neuro-symbolic friction learning module and a physics-informed navigation module.} The neuro-symbolic module has a neural network to predict friction coefficients from visual inputs, guided by symbolic physics reasoning for self-supervised training. 
    The inputs are captured by camera and LiDAR before entering a terrain patch, while training occurs afterward using the dynamics observed on that patch.
    The navigation module builds terrain property maps online and leverages the predicted friction information to generate physically feasible and efficient path and speed profiles for off-road driving.}
    \label{fig:NSL_flow}
\end{figure*}

\subsection{Data-driven Approaches}

To address the limitations of physics-based methods, data-driven approaches for terrain analysis have gained significant attention. For instance, \cite{vulpi2021recurrent, ewen2022these} employed neural networks to predict terrain properties through visual segmentation. \cite{wellhausen2019should, castro2023does} estimated terrain traversability using motion data derived from torque sensors or IMU measurements. \cite{jeon2024follow} proposed to estimate off-road traversability through wheel track demonstrations during pre-driving. \cite{gasparino2024wayfaster, meng2023terrainnet} trained a model to predict overlooking traversability maps directly from front-view camera inputs. These methods exhibit notable advancements in capturing the variability and complexity inherent in unstructured off-road environments.

In addition to terrain analysis, machine learning techniques have been applied to predicting robot dynamics. \cite{narayanan2020gated, tremblay2021multimodal, wang2023sa} utilized neural networks to effectively capture sequential dependencies, delivering robust state and trajectory predictions over extended time horizons. 
\cite{xiao2024anycar} developed a transformer-based vehicle dynamics model using extensive training data from multiple simulators.
Besides, reinforcement learning techniques have also been adapted to address off-road driving in an end-to-end manner, leveraging experiences the agent learned in simulated environments \citep{wang2022end, wang2023deep, wang2024pay}.
However, these data-driven approaches face challenges in generalization, largely due to their dependence on extensive labeled datasets and the absence of explicit physical reasoning.
Benchmark datasets such as TartanDrive \citep{triest2022tartandrive} have provided valuable resources for studying off-road robot dynamics across diverse terrain types. However, its associated dynamics model is primarily data-driven and does not explicitly incorporate physical constraints.

\subsection{Physics-Informed Networks}

The fusion of physics-based insights with neural networks has emerged as a promising direction for modeling dynamic systems. Approaches such as Neural Ordinary Differential Equations \citep{chen2018neural} and Hamiltonian Neural Networks \citep{greydanus2019hamiltonian} embed conservation laws and system dynamics into learning processes, preserving the structure of physical systems. Variational Integrator Networks \citep{saemundsson2020variational} and their extensions \citep{havens2021forced, duruisseaux2023lie} have shown promise in imposing geometric and physical constraints, such as symmetries. These methods have demonstrated accuracy in modeling simple physics, such as pendulums and mass-spring systems. However, real-world applications remain challenging due to the increased complexity and uncertainty inherent in practical environments, as highlighted by \cite{zhao2024physord}.

Recent advancements in physics-informed learning have demonstrated significant potential in addressing such challenges, particularly in off-road applications.
\cite{frey2024roadrunner} proposed a self-supervised framework that trains a traversability prediction network using pseudo-labels generated by X-Racer, a physics-based off-road autonomy stack. \cite{zhao2024physord} integrated learnable components into the vehicle's Lagrangian equations, constraining the learning process with physics principles such as energy conservation. \cite{agishev2023monoforce} predicted terrain height and stiffness properties from visual inputs, integrating the calculated supporting force through a differentiable physics engine for supervision. \cite{chen2024identifying} introduced a network for estimating friction and stiffness from proprioceptive observations, initially trained in simulation and later adapted to real-world environments. \cite{pokhrel2024cahsor, datar2024terrain, datar2024learning} trained a neural kinodynamic model using aggressive driving data and employed it as a safety constraint for autonomous driving.
\cite{cai2024pietra, cai2023probabilistic} learns empirical traction distributions and uncertainty with evidential deep learning, and plans navigation paths according to worst‑case traction costs.
However, accurately predicting the friction coefficient of diverse off-road terrains and leveraging this knowledge for reliable long-term navigation remains a challenging open problem. 


\section{Neuro-Symbolic Friction Learning} \label{sec:NSL}

\paragraph{Overview}

We propose the AnyNav framework, illustrated in \fref{fig:NSL_flow}, which consists of two primary modules: a neuro-symbolic module for friction coefficient estimation with its self-supervised training strategy described in \sref{sec:NSL}, and a physics-informed navigation system leveraging the predicted frictions, presented in \sref{sec:planner}.
This architecture decouples terrain-specific physics modeling from vehicle-related navigation tasks, enhancing interpretability and generalizability.
The system takes LiDAR, camera, IMU, and wheel revolutions-per-minute (RPM) sensor data as inputs, and produces friction coefficient estimates along with path and speed plans for navigation.

\subsection{Neural Perception Module} \label{sec:neural_perception}

Instead of directly predicting frictional forces, which depend on instantaneous motion states and control inputs, our approach focuses on estimating the terrain-specific friction coefficients that remain invariant to vehicle dynamics. These coefficients describe the intrinsic contact properties of a terrain and can be inferred from its visual and geometric appearance.
Our perception module leverages a neural network to process visual inputs and regress continuous coefficients that characterize the frictional behavior of each terrain pixel.
By predicting continuous coefficients rather than discrete terrain categories as in prior works \citep{vulpi2021recurrent, ewen2022these}, our model can generalize beyond a fixed set of terrain types in real-world conditions.

\begin{figure}
    \centering
    \begin{subfigure}[h]{0.8\linewidth}
        \includegraphics[width=\textwidth]{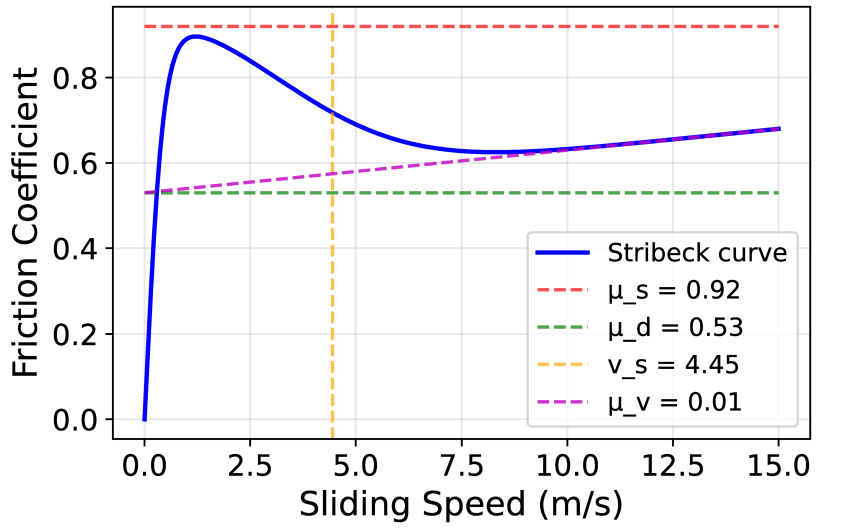}
    \end{subfigure}
    \vspace{-2.3mm}
    \caption{\textbf{An example of the Stribeck curve}. The plot illustrates the friction coefficient of a terrain material with respect to tire sliding speed. The shape of the curve is governed by four coefficients, which vary across different terrain types.}
    \label{fig:example_stribeck}
\end{figure}

To enable real-time operation, we accumulate and project online LiDAR and camera data to generate local bird’s-eye view (BEV) images and height patches, which serve as the four-channel network inputs $\mathbf{X}_{\text{in}}$, following the method of \cite{triest2022tartandrive}.
The outputs $\mathbf{S}_{\text{pred}}$ are defined in the Stribeck model space \citep{armstrong1995canudas} to characterize the varying frictional conditions.
The Stribeck model is widely regarded as more accurate than the conventional Coulomb model \citep{popov2017coulomb} because it regards the transition from static to dynamic friction (the Stribeck effect) and incorporates viscous damping.
In the Stribeck model, the effective friction coefficient $\mu$ can be computed as a function of the wheel skidding speed $v_{\text{rel}}$ (relative speed between tire and ground), parameterized by four terrain-related coefficients $\{\mu_s, \mu_d, v_S, \mu_v\}$:
\begin{equation} \label{eq:stribeck}
\begin{aligned}
    \mu =& \text{Stribeck}(v_{\text{rel}}; \mu_s, \mu_d, v_S, \mu_v) \\
    =& \sqrt{2e}(\mu_s-\mu_d)\exp\left(-\left(\frac{v_{\text{rel}}}{v_S}\right)^2\right)\frac{v_{\text{rel}}}{v_S} \\
    &+ \mu_d\tanh\left(\frac{10\sqrt{2}v_{\text{rel}}}{v_S}\right) + \mu_v v_{\text{rel}},
\end{aligned}
\end{equation}
where $\mu_s$ represents the static friction coefficient before the wheel begins to slide, $\mu_d$ denotes the dynamic friction coefficient for full sliding, $v_S$ defines the threshold sliding speed marking the transition from static to dynamic friction, and $\mu_v$ characterizes the viscous resistance. 
An example of a Stribeck curve is illustrated in \fref{fig:example_stribeck}, which depicts the relationship between the effective friction coefficient and tire sliding velocity.
Typically, each terrain type corresponds to a distinct set of Stribeck coefficients. The neural network predicts Stribeck coefficients $\mathbf{S}_{\text{pred}} \triangleq \{\mu_s, \mu_d, v_S, \mu_v\}$ based on the visual and geometrical features. 

For efficiency, we adopt the lightweight UNet \citep{ronneberger2015u} as the perception network. To accommodate the data dimension compatibility, we modify the first and last convolutional layers to 4 channels. As the Stribeck coefficients are positive by their physical definition, we attach a rectification layer to the UNet:
\begin{equation} 
    \mathbf{S}_{\text{pred}} = f_{\boldsymbol\theta}(\mathbf{X}_{\text{in}}) = \mathbf{A} \cdot \text{sigmoid}(\text{UNet}_{\boldsymbol\theta}(\mathbf{X}_{\text{in}})), 
\end{equation}
where the $\text{sigmoid}(\cdot)$ function maps the unbounded network output to the range $(0,1)$, and each channel is subsequently scaled by $\mathbf{A} = [1.0\;\;1.0\;\;10\;\;0.02]$. 
This constrains the predicted Stribeck coefficients within their reasonable ranges, thereby avoiding invalid numerical issues in the subsequent physics module and ensuring stable convergence. The ranges are summarized from literature studying tire dynamics and friction curves \citep{holloway1989examination, canudas2003dynamic, anderson2015interaction}.
The network is applied independently to each wheel, and its model weights are shared across all wheels under the assumption of identical structure and dynamics. This further ensures that the model captures general physical principles applicable to all wheels.

\subsection{Symbolic Physics Module} \label{sec:symbolic_reasoning}

The symbolic physics module computes vehicle dynamics based on the predicted friction coefficients, terrain geometry, and vehicle model. The physical rules applied in this module serve to guide model training, which will be detailed in \sref{sec:training}.
Here, we first introduce the vehicle modeling in \sref{ssec:weight-modeling}, perform force analysis in \sref{ssec:force-analysis}, and compute the inferred motion in \sref{ssec:integrate-semi}.

\begin{figure}
    \centering
    \begin{subfigure}[b]{0.8\linewidth}
        \includegraphics[width=\textwidth]{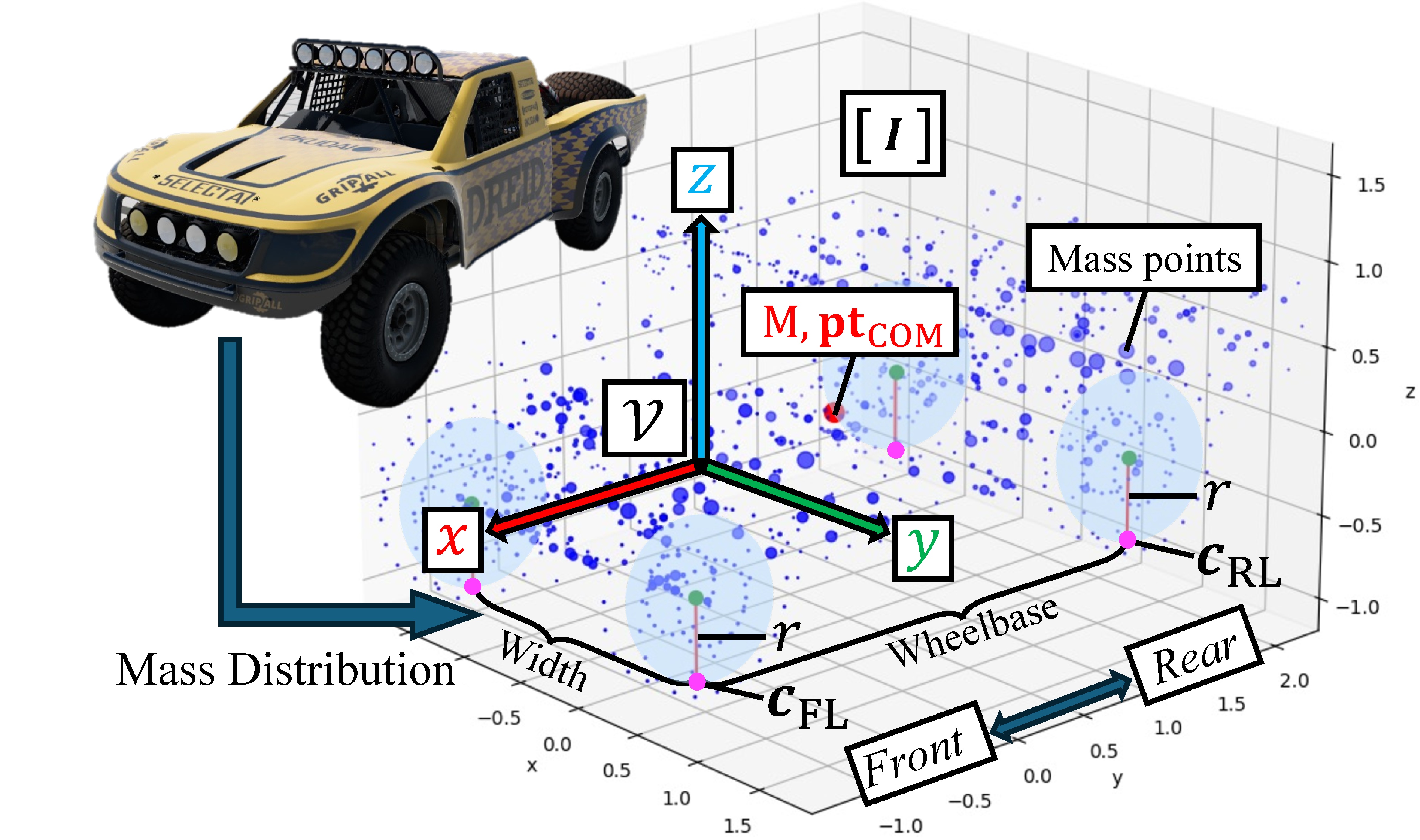}
    \end{subfigure}
    \caption{\textbf{The configurable vehicle model}. The blue dots represent the mass distribution, with their size proportional to the magnitude of the mass at each point. $M$ is the total mass, $\mathbf{pt}_{\text{COM}}$ is the center of mass, $\mathbf{I}$ is the rotational inertia matrix, $\mathbf{c}$ are wheel contact points, and $r$ is wheel radius.}
    \label{fig:mass}
\end{figure}

\subsubsection{Vehicle Model}
\label{ssec:weight-modeling}

To be generalizable, we represent the vehicle as a rigid body fixed in a local coordinate frame $\mathcal{V}$.
Denote the vehicle's total mass as $M$, the center of mass (COM) as $\textbf{pt}_{\text{COM}}^{\mathcal{V}} \in \mathbb{R}^3$, and the rotational inertia matrix around the COM as $\textbf{I}^{\mathcal{V}} \in \mathbb{R}^{3\times3}$.
When the exact mass distribution is known, such as in simulation, the inertia matrix can be computed following \citep{witkin1997physically}:
\begin{equation}
\begin{aligned}
    \boldsymbol{\delta}_i &= \textbf{pt}_i^{\mathcal{V}} - \textbf{pt}_{\text{COM}}^{\mathcal{V}}, \\
    \textbf{I}^{\mathcal{V}} &= \sum_{i=1}^{N_m} \text{mass}_i \left( \lVert \boldsymbol{\delta}_i \rVert^2 \textbf{I}_{3\times 3} - \boldsymbol{\delta}_i \boldsymbol{\delta}_i^{T} \right),
\end{aligned}
\end{equation}
where $\{(\textbf{pt}_i^{\mathcal{V}}, \text{mass}_i)\}$ denote the positions and masses of the $N_m$ mass points.
When the exact model of the vehicle is unknown, such as in real-world testing, several approaches exist to estimate the mass distribution and inertia matrix, such as \citep{bottelli2014online}.
Our experiments show that assuming a uniform mass distribution is sufficient in real-world testing, since the vehicle model is jointly optimized in our bilevel optimization illustrated in \sref{sec:finetuning_real}, and thus it has minimal impact on the performance.
We also model the geometric properties of the vehicle by defining the wheel radius $r_w \in \mathbb{R}^+$ and the ground contact points of each wheel $\mathbf{c}_w^{\mathcal{V}} \in \mathbb{R}^3$, where $w \in \{\text{FL, FR, RL, RR}\}$ denotes front-left, front-right, rear-left, and rear-right, respectively.
An exemplar vehicle model is illustrated in \fref{fig:mass}.

\subsubsection{Force Analysis} \label{ssec:force-analysis}

To derive the vehicle dynamics, we analyze the forces acting on each wheel using a physics-based model. Several factors are considered, including the vehicle’s current state, ground slope, predicted friction coefficient, and tire slip rates, to compute the normal support forces and lateral friction forces.
The normal forces acting at each tire contact point are perpendicular to the ground plane, supporting the vehicle against gravity. These forces are not evenly distributed across all four wheels during acceleration or driving on a slope. According to the weight transfer theory \citep{su2023double}, the ratio of front-to-rear normal force allocation can be computed as
\begin{equation}
\beta \triangleq \frac{F_N^{\text{front}}}{F_N^{\text{rear}}} = \frac{- hMg\sin\theta - hMa + d_r Mg\cos\theta}{hMg\sin\theta + hMa + d_f Mg\cos\theta},
\end{equation}
where $h$ is the height of the COM w.r.t. the ground plane, $d_f$ and $d_r$ are the distances along the slope from the COM to the front and rear contact points, and $\theta$ is the slope angle.
The detailed derivation process is provided in \aref{ssec:weight-transfer}.
A similar process can be applied to compute the transverse force ratio $\gamma \triangleq F_N^{\text{left}}/F_N^{\text{right}}$. 
Therefore, the magnitudes of individual normal forces at four wheels are
\begin{equation} \label{eq:normal_mag}
\begin{aligned}
    F_N^{\text{FL}} &= \frac{\beta}{1+\beta} \cdot \frac{\gamma}{1+\gamma} \cdot Mg\cos\theta, \\
    F_N^{\text{FR}} &= \frac{\beta}{1+\beta} \cdot \frac{1}{1+\gamma} \cdot Mg\cos\theta, \\
    F_N^{\text{RL}} &= \frac{1}{1+\beta} \cdot \frac{\gamma}{1+\gamma} \cdot Mg\cos\theta, \\
    F_N^{\text{RR}} &= \frac{1}{1+\beta} \cdot \frac{1}{1+\gamma} \cdot Mg\cos\theta.
\end{aligned}
\end{equation}
Finally, the normal force vectors $\mathbf{F}_N^w$ can be obtained by multiplying the magnitudes computed in \eqref{eq:normal_mag} by a unit normal vector pointing in the direction perpendicular to the slope.

The friction force depends on the predicted friction coefficients $\mathbf{S}_{\text{pred}}^w$ and the skidding speed of the wheel $\mathbf{v}_{\text{rel}}^w$. 
The skidding speed is non-zero if the wheel is losing traction on the ground. It is measured by the difference between the linear velocity of the wheel's center and its edge,
\begin{equation}
    \mathbf{v}_{\text{rel}}^w = \mathbf{v}_e^w - \mathbf{v}_c^w.
\end{equation}
where the wheel’s linear velocity $\mathbf{v}_c^w$ is computed from the vehicle body’s linear and angular velocities, considering the wheel's offset from the COM; the wheel edge velocity $\mathbf{v}_e^w$ is determined from the wheel’s RPM measurement and radius.
We sample the Stribeck curve at the calculated skidding speed $\mathbf{v}_{\text{rel}}^w$ to determine the friction coefficient of each wheel,
\begin{equation}
    \mu^w = \text{Stribeck}(||\mathbf{v}_{\text{rel}}^w||; \mathbf{S}_{\text{pred}}^w).
\end{equation}
Combining with the estimated normal force $\mathbf{F}_N^w$ allows us to calculate the friction force exerted on a wheel $w$
\begin{equation}
    \mathbf{F}_f^w = -\mu^w||\mathbf{F}_N^w||\cdot \frac{\mathbf{v}_{\text{rel}}^w}{||\mathbf{v}_{\text{rel}}^w||},
\end{equation}
pointing in the opposite direction of the relative velocity.

\subsubsection{Semi-implicit Euler Integrator}
\label{ssec:integrate-semi}

We then integrate the calculated forces to predict the vehicle’s trajectory. The semi-implicit Euler method \citep{liu2004convergence} is employed for integration, as it provides a good balance between numerical stability and computational efficiency.
It estimates the vehicle's motion considering its pose $\boldsymbol\xi=[\mathbf{t} \;\; \mathbf{q}]$, linear velocity $\mathbf{v}$, and angular velocity $\boldsymbol\omega$, from an initial state $\{\boldsymbol\xi_0, \mathbf{v}_0, \boldsymbol\omega_0\}$ using the vehicle's acceleration. The total external force and resulting torque are
\begin{equation}
    \mathbf{F} = M g + \sum_w (\mathbf{F}_N^w + \mathbf{F}_f^w),
\end{equation}
\begin{equation}
    \boldsymbol\tau = \sum_w (\mathbf{c}_w - \mathbf{pt}_{\text{COM}}) \times (\mathbf{F}_N^w + \mathbf{F}_f^w).
\end{equation}
The linear and angular accelerations are computed as
\begin{equation}
    \mathbf{a} = \frac{d\mathbf{v}}{dt} = \frac{\mathbf{F}}{M},
    \label{eq:accel}
\end{equation}
\begin{equation}
    \boldsymbol\alpha = \frac{d\boldsymbol\omega}{dt} = (\mathbf{I}^\mathcal{W})^{-1} \boldsymbol\tau,
    \label{eq:ang_accel}
\end{equation}
where $\mathbf{I}^\mathcal{W} = \mathbf{q}^{-1} \mathbf{I}^\mathcal{V} \mathbf{q}$ is the inertia matrix in world coordinate.
Finally, we apply the semi-implicit Euler method to integrate the velocity, angular velocity, and pose over a time step $\Delta t$,
\begin{equation}
    \mathbf{v}_{\Delta t} = \mathbf{v} + \mathbf{a}\Delta t,
\end{equation}
\begin{equation}
    \boldsymbol\omega_{\Delta t} = \boldsymbol\omega + \boldsymbol\alpha\Delta t,
\end{equation}
\begin{equation}
    \mathbf{t}_{\Delta t} = \mathbf{t} + \mathbf{v}_{\Delta t}\Delta t,
\end{equation}
\begin{equation}
    \mathbf{q}_{\Delta t} = \text{Normalize}\left(\mathbf{q} + \frac{1}{2} \left([\boldsymbol\omega_{\Delta t}, 0] \mathbf{q}\right) \Delta t\right),
\end{equation}
where $[\boldsymbol\omega_{\Delta t}, 0]$ is a quaternion with an imaginary part of $\boldsymbol\omega_{\Delta t}$ and a real part of 0. $\text{Normalize}(\cdot)$ projects the computed result back to a unit quaternion.
By performing the integration iteratively over time, we can estimate the trajectory over $N$ steps.

\subsection{Training Strategies} \label{sec:training}

One major challenge in predicting friction coefficients is obtaining physically realistic ground truth.
This is particularly difficult because the friction coefficient is a latent variable within the vehicle-terrain interaction, making it impossible to measure directly without specialized equipment. To address this, we design self-supervised training strategies that infer frictional properties through physical reasoning, thereby grounding the neural model in physical laws. Depending on data availability and noise level, we introduce a training method for simulation and a sim-to-real transfer approach for real-world scenarios.

\subsubsection{Training in Simulation} \label{sec:training_sim}

Simulated environments offer a unique advantage for training neural networks from scratch. Unlike the real world, these environments simplify physical effects and provide precise, noiseless sensor measurements, making them ideal for initial training. 
Our neuro-symbolic learning approach combines two loss functions: an acceleration loss and a prior knowledge loss.

The acceleration loss function is designed to align the inferred and observed motions. Since the inferred motion is computed based on the predicted friction coefficients, better consistency with the actual motion indicates more accurate friction prediction.
The formulation is given by
\begin{equation}
    L_{\text{acc}} = \text{Huber}(\mathbf{a}_{\text{est}}, \mathbf{a}_{\text{meas}}) + \text{Huber}(\boldsymbol\alpha_{\text{est}}, \boldsymbol\alpha_{\text{meas}}),
\end{equation}
where $\mathbf{a}_{\text{est}}, \boldsymbol\alpha_{\text{est}}$ are the estimated linear and angular accelerations defined in \eqref{eq:accel}-\eqref{eq:ang_accel}, $\mathbf{a}_{\text{meas}}, \boldsymbol\alpha_{\text{meas}}$ denote their measurements obtained from the simulation.
The Huber kernel \citep{huber1992robust} is employed to mitigate the influence of outlier data samples, such as cases where the tires lose ground contact when driving over rough terrain.
During training, the gradients of this loss are backpropagated through the differentiable physics module to update the neural network parameters.

While the acceleration loss applies to the inferred kinematics, the prior-knowledge loss, derived from accurate simulated terrain labels, directly applies to the Stribeck coefficients. Specifically, for each terrain type $u$, we assign a prior $\mathbf{S}_{\text{prior}}(u)$ to represent its typical values of Stribeck coefficients. The prior knowledge loss is
\begin{equation}
    L_{\text{prior}} = \text{L1Loss}(\mathbf{S}_{\text{pred}}, \mathbf{S}_{\text{prior}}(u)).
\end{equation}
The final loss function is a combination of the two,
\begin{equation} \label{eq:sim_tot_loss}
    L = L_{\text{acc}} + \lambda L_{\text{prior}},
\end{equation}
This ensures the predicted friction effects align with both the observed motion and the typical terrain property. 

\subsubsection{Self-Supervised Sim-to-Real Transfer} \label{sec:finetuning_real}

Training in real-world settings presents greater challenges compared to simulation environments. Two major obstacles arise: labeling terrain types is difficult, and noisy sensor measurements introduce significant bias into the physics model. 
Additionally, substantial uncertainties exist in the vehicle model, particularly in determining the mass distribution and inertia matrix. To address these challenges, we develop a self-supervised approach based on bilevel optimization, inspired by imperative learning \citep{wang2025imperative}.
This approach jointly refines the predicted Stribeck coefficients, sensor noises, and the vehicle's inertia matrix to satisfy a set of physical constraints, thereby mitigating the effects of sensor noise and vehicle model uncertainty.
Specifically, we solve the problem \eqref{eq:bilevel} with alternating optimization, where the lower-level physics-based optimization \eqref{eq:bilevel-lower} and the upper-level network optimization \eqref{eq:bilevel-upper} are performed iteratively until convergence. This results in self-supervised learning by jointly enforcing consistency between the predicted terrain properties and the observed vehicle dynamics.
\begin{subequations}\label{eq:bilevel}
\begin{align}
    &\min_{\boldsymbol\theta} \;\; ||f_{\boldsymbol\theta}(\mathbf{X}_{\text{in}}) - \mathbf{S}^*||_1 \label{eq:bilevel-upper} \\
    &~\text{s.t. \;}\mathbf{S}^* = \arg\min_{\mathbf{S}} \;\; C(\mathbf{S}_{\text{pred}}).
    \label{eq:bilevel-lower}
\end{align}
\end{subequations}
The lower-level optimization \eqref{eq:bilevel-lower} is the key to ensuring the entire physics-grounded learning procedure is self-supervised. Since the friction coefficient is difficult to observe directly, we employ several indirect constraints to align the inferred and measured motions that are affected by friction coefficients, ensuring both physical and temporal consistency.
In addition to the friction coefficients, we also jointly optimize the vehicle inertia matrix and individual wheel speeds. The inertia matrix is included to compensate for modeling inaccuracies arising from unknown mass distributions, while wheel speed is treated as an optimization variable to improve robustness against measurement noise.
Specifically, the lower-level optimization consists of four indirect physical constraints conditioning on the estimated Stribeck coefficients $\mathbf{S}_i$, the vehicle's inertia matrix $\mathbf{I}^\mathcal{V}$, and wheel speed $\text{RPM}_i$ for a time horizon $i=1 \cdots N$, where $N$ is the number of time steps.

\myparagraph{Dynamics Constraints} enforce the accelerations estimated by the physics module \eqref{eq:accel}-\eqref{eq:ang_accel} align with the measurements. These measurements can be obtained from a calibrated IMU or a location and mapping system.
\begin{equation}
    \mathbf{C}_a^{(i)} = \mathbf{a}_{\text{est}}^{(i)} - \mathbf{q}_i\mathbf{a}_{\text{meas}}^{(i)},
\end{equation}
\begin{equation}
    \mathbf{C}_\alpha^{(i)} = \boldsymbol\alpha_{\text{est}}^{(i)} - \mathbf{q}_i\boldsymbol\alpha_{\text{meas}}^{(i)}.
\end{equation}

\myparagraph{Integration Constraints} ensure the rotation and velocity are consistent with acceleration integration over $i$ to $i+1$,
\begin{equation}
    \mathbf{C}_v^{(i)} = (\mathbf{v}_{i+1} - \mathbf{v}_i) - \mathbf{q}_i\mathbf{a}_{\text{meas}}^{(i)} \Delta t,
\end{equation}
\begin{equation}
    \mathbf{C}_q^{(i)} = \text{Log}\left(\mathbf{q}_{i+1}^{-1} \left(\mathbf{q}_i + \frac{1}{2} \left([\boldsymbol\omega_{\text{meas}}^{(i)}, 0]\mathbf{q}_i\right) \Delta t\right)\right),
\end{equation}
where $\text{Log}(\cdot)$ is the Log mapping to Lie algebra.

\myparagraph{Stribeck Constraint} ensures that the Stribeck coefficients remain within reasonable ranges and avoid rapid fluctuations, thereby promoting temporal consistency,
\begin{equation}
    \mathbf{C}_s^{(i)} = \max\left(0, \frac{\mathbf{S}_i}{\mathbf{A}}-1\right)^2 + \left(\mathbf{S}_{i+1} - \mathbf{S}_i\right),
\end{equation}
where $A$ denotes the channel-wise upper bounds of the Stribeck coefficients, as defined in \sref{sec:neural_perception}.

\begin{algorithm}[t]
\caption{Alternating Bilevel Optimization}
\label{alg:bilevel}
\begin{algorithmic}[1]
\Function{LowerLevel}{$\mathbf{S}, \mathbf{I}^{\mathcal V}, \mathrm{RPM}, \mathbf{a}_{\text{meas}}, \boldsymbol\alpha_{\text{meas}}$}
    \While{\textbf{not} LMOptimizer.Converged()}
        \State $\mathbf{a}_{\text{est}}, \boldsymbol\alpha_{\text{est}}, \mathbf{v}_{\text{est}}, \mathbf{q}_{\text{est}} \gets \text{PhysMdl}(\mathbf{S}, \mathbf{I}^{\mathcal V}, \mathrm{RPM})$ \\
        \Comment{Forward physics module (\sref{sec:symbolic_reasoning})}
        \State $C \gets \text{Cost}(\mathbf{a}_{\text{est}}, \boldsymbol\alpha_{\text{est}}, \mathbf{v}_{\text{est}}, \mathbf{q}_{\text{est}}, \mathbf{a}_{\text{meas}}, \boldsymbol\alpha_{\text{meas}})$ \\
        \Comment{Lower-level objective (\sref{sec:finetuning_real}, \eqref{eq:tot_il_cost})}
        \State $\mathbf{S}, \mathbf{I}^{\mathcal V}, \mathrm{RPM} \gets \text{LMOptimizer.Step}(C)$
    \EndWhile
    \State \Return $\mathbf{S}$
\EndFunction
\\
\Function{UpperLevel}{$\text{Dataset}, \mathbf{I}^{\mathcal V}$}
    \For{$\text{epoch} \in 1 \cdots \text{max\_epoch}$}
    \ForAll{$\text{batch} \in \text{Dataset}$}
        \State $\mathbf{X}_{\text{in}}, \mathrm{RPM}, \mathbf{a}_{\text{meas}}, \boldsymbol\alpha_{\text{meas}} \gets \text{Unpack}(\text{batch})$
        \State $\mathbf{S}_{\text{pred}} \gets f_{\boldsymbol\theta}(\mathbf{X}_{\text{in}})$ \\
        \Comment{Forward friction network (\sref{sec:neural_perception})}
        \State $\mathbf{S}^{*} \gets \Call{LowerLevel}{\mathbf{S}_{\text{pred}}, \mathbf{I}^{\mathcal V}, \mathrm{RPM}, \mathbf{a}_{\text{meas}}, \boldsymbol\alpha_{\text{meas}}}$
        \State $L \gets \text{L1Loss}(\mathbf{S}_{\text{pred}}, \mathbf{S}^{*})$
        \State $\boldsymbol\theta \gets \text{AdamOptimizer.Step}(L)$
    \EndFor
    \EndFor
\EndFunction
\end{algorithmic}
\end{algorithm}

\myparagraph{Wheel Speed Constraint} ensures that the wheel revolutions-per-minute (RPM) variable remains close to its measurement,
\begin{equation}
    \mathbf{C}_w^{(i)} = \text{RPM}_i - \text{RPM}_{\text{meas}}^{(i)}.
\end{equation}
The objective of lower-level optimization is the weighted sum of all these constraints across all time steps,
\begin{equation}\label{eq:tot_il_cost}
\scalebox{0.9}{$
    C = \sum_{i=1}^{N-1} \left(\begin{matrix}
        \mathbf{C}_a^{(i)T}\mathbf{W}_a\mathbf{C}_a^{(i)} + \mathbf{C}_\alpha^{(i)T}\mathbf{W}_\alpha\mathbf{C}_\alpha^{(i)} + \mathbf{C}_v^{(i)T}\mathbf{W}_v\mathbf{C}_v^{(i)} \\
        + \mathbf{C}_q^{(i)T}\mathbf{W}_q\mathbf{C}_q^{(i)} + \mathbf{C}_s^{(i)T}\mathbf{W}_s\mathbf{C}_s^{(i)} + \mathbf{C}_w^{(i)T}\mathbf{W}_w\mathbf{C}_w^{(i)}
    \end{matrix}\right),
$}
\end{equation}
where $\mathbf{W}_{(\cdot)}$ are diagonal weighting matrices used to balance each term.
In practice, the Stribeck coefficients in the lower-level optimization are initialized using predictions from the network, $\mathbf{S}_i = f_{\boldsymbol\theta}(\mathbf{X}_{\mathrm{in}}^{(i)})$. These predictions serve as numerical initializations and are subsequently refined through physics-based optimization. The inertia matrix $\mathbf{I}^{\mathcal{V}}$ is initialized by assuming a uniform mass distribution, while $\mathrm{RPM}_i$ is initialized using the measured wheel speed.
The second-order Levenberg-Marquardt (LM) algorithm in PyPose \citep{wang2023pypose} is employed to iteratively solve the lower-level optimization problem.
The complete procedure of the alternating bilevel optimization is detailed in \algorithmref{alg:bilevel}.

\section{Physics-Informed Navigation System} \label{sec:planner}

Our physics-informed navigation system builds on the capability of friction estimation. By understanding this important terrain property, the system can make informed decisions about both route selection and speed planning, taking into account the physical terrain-vehicle interactions.
For example, a ``smart'' navigator may prefer high-friction areas to facilitate hill climbing and slow down when navigating low-friction areas to ensure stable steering. 
The system can operate on a prebuilt global map constructed from drone scans or previous vehicle traversals, or use onboard sensors to build a real-time physical map of a new environment for navigation.
We next present the design of this physics-informed navigation system, including the system architecture in \sref{sec:navigation_overview}, the mapping system in \sref{sec:nav_map}, the path planner in \sref{sec:path_planner}, and the speed planner in \sref{sec:speed_planner}.

\begin{figure}[!t]
    \centering
    \includegraphics[width=\linewidth]{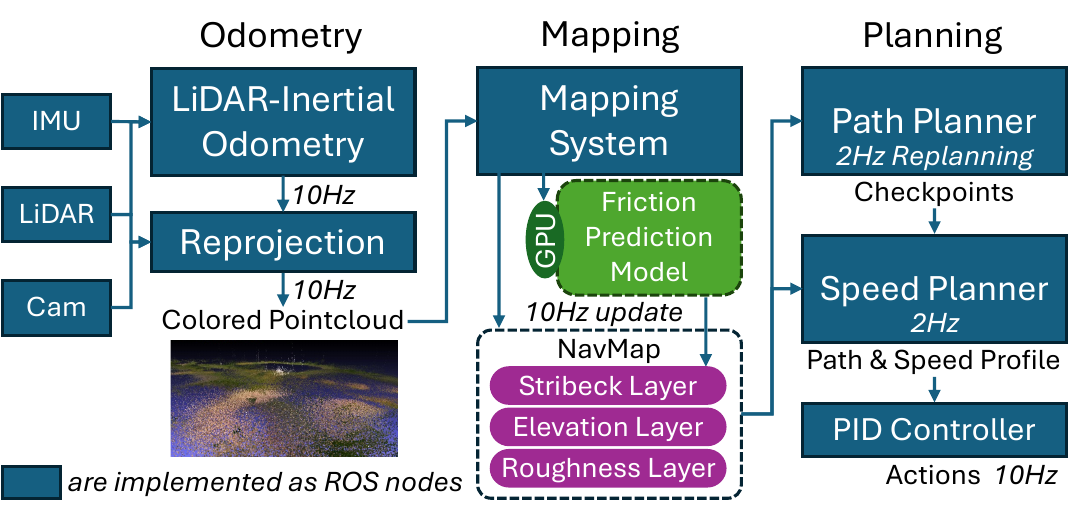}
    \caption{\textbf{Architecture of the AnyNav system.} Blue boxes represent ROS nodes; each operates in parallel processes. The odometry and mapping stacks run at 10 Hz, while the planning stack runs at 2 Hz for path and speed planning and at 10 Hz for PID control.}
    \label{fig:impl}
\end{figure}

\subsection{System Architecture}
\label{sec:navigation_overview}

To ensure real-time performance, our system adopts a multiprocessing architecture in Robot Operating System (ROS) \citep{quigley2009ros}, as illustrated in \fref{fig:impl}. The odometry nodes run SuperOdometry \citep{zhao2025resilient} at 10 Hz, matching the sensor update rate, to provide vehicle poses and colored point clouds by fusing LiDAR scans and camera images. The mapping module also operates at 10 Hz, updating the terrain property map using the current frame, with a dedicated GPU thread for friction estimation. The path and speed planners run at 2 Hz, continuously replanning on the latest terrain map to generate feasible and efficient path and speed profiles. Finally, the PID controller node runs at 10 Hz, producing commands executed by the vehicle.

\subsection{Mapping System} \label{sec:nav_map}

We introduce a physically grounded representation of the environment that captures terrain properties critical for off-road navigation. Our mapping system maintains this terrain property map dynamically, enabling real-time updating and expansion as new regions are observed.
The map is represented as a 2.5D grid composed of three layers: the Stribeck layer $\mathbf{SL}$ encodes predicted friction coefficients; the elevation layer $\mathbf{EL}$ represents ground height; and the roughness layer $\mathbf{RL}$ represents the terrain bumpiness. Specifically, for each grid cell $(x,y)$, we have
\begin{subequations}
\label{eq:map_update}
\begin{align}
    \mathbf{SL}_{\text{xy}} &= f_{\boldsymbol\theta}(\mathbf{X}^{\text{in}}_{\text{xy}}), \label{eq:SL_layer} \\
    \mathbf{EL}_{\text{xy}} &= \text{Mean}(z \mid z \in \mathbf{Z}_{\text{xy}},\, z < h + \text{Min}(\mathbf{Z}_{\text{xy}})), \label{eq:EL_layer} \\
    \mathbf{RL}_{\text{xy}} &= \text{Var}(z \mid z \in \mathbf{Z}_{\text{xy}},\, z < h + \text{Min}(\mathbf{Z}_{\text{xy}})), \label{eq:RL_layer}
\end{align}
\end{subequations}
where the BEV terrain patch $\mathbf{X}^{\text{in}}_{\text{xy}}$ and heights $\mathbf{Z}_{\text{xy}}$ of grid cell $(x,y)$ are incrementally constructed from the sensor data stream, and $h$ defines a height threshold above the ground for removing overhanging objects such as tree crowns.

\begin{figure}[t]
    \centering
    \includegraphics[width=\linewidth]{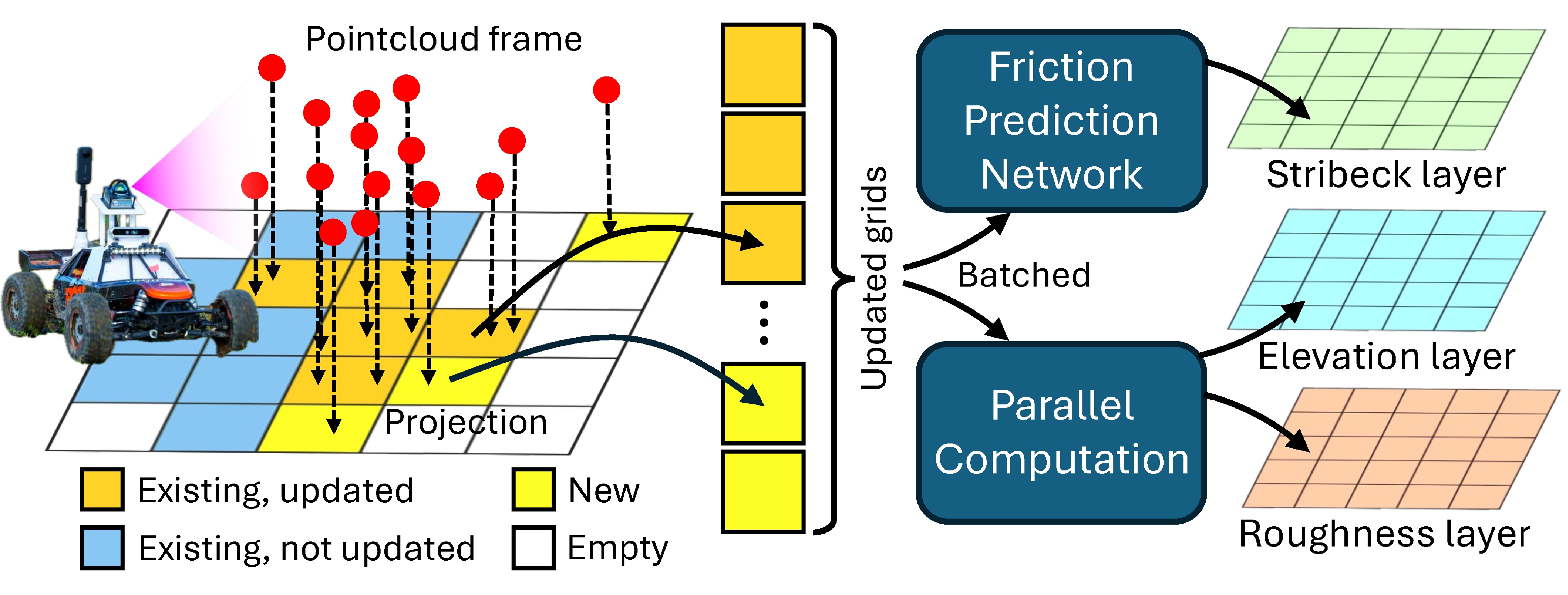}
    \caption{\textbf{Illustration of the real-time mapping process.} For each frame, the colorized points are projected onto the map grid. New grid cells are instantiated as needed. The updated cells are then used to update the three terrain property layers.}
    \label{fig:map_proc}
\end{figure}

The map expands in real time as the robot explores new areas, as illustrated in \fref{fig:map_proc}. We project each incoming colorized point cloud onto grid cells to update their internal buffers of $\mathbf{X}^{\text{in}}_{\text{xy}}$ and $\mathbf{Z}_{\text{xy}}$ based on the estimated robot poses.
For faster processing, cells are instantiated only when they are observed, and later updated in batches per \eqref{eq:map_update}.
When memory space becomes limited, cells far from the robot’s current position are offloaded to hard disk and reloaded on demand, resulting in a memory-efficient mapping system.

\begin{figure}[t]
    \centering
    \includegraphics[width=\linewidth]{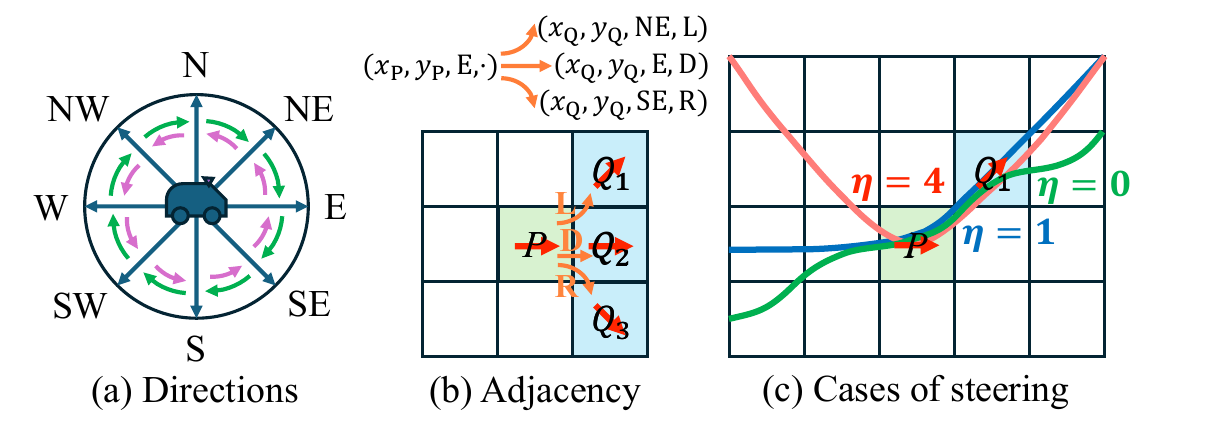}
    \caption{ 
    (a) The eight discretized orientations of the vehicle.
    (b) Example of graph edges: vertex P (east-facing) connects to $\text{Q}_1$, $\text{Q}_2$, and $\text{Q}_3$ via left-turn, straight, and right-turn actions, respectively.
    (c) Depiction of the turning curvature factor: $\eta = 4$ for consecutive turns (e.g., $m_\text{P} = m_\text{Q} = \text{L}$), $\eta = 1$ for a single turning action (e.g., $m_\text{P} = \text{D}, m_\text{Q} = \text{L}$), and $\eta = 0$ for opposite turns (e.g., $m_\text{P} = \text{R}, m_\text{Q} = \text{L}$) that cancel after smoothing.
    }
    \label{fig:steering}
\end{figure}

\subsection{Path Planner} \label{sec:path_planner}
To ensure efficiency in long-range navigation, which is essential for fully utilizing the map information to proactively avoid hazards far ahead, we separate path and speed planning into two consecutive processes. This separation substantially reduces the complexity of each subproblem, making real-time, long-range planning feasible on onboard devices. 
In practice, this approach yields reliable navigation plans without path and speed issues.
We next introduce the path planner and the speed planner separately.

To generate physically feasible and safe routes, we incorporate both terrain properties and vehicle status into path planning. 
Specifically, we introduce a heading-steering-aware graph for the A* search algorithm \citep{hart1968formal} to integrate vehicle heading and steering information, where each vertex $(x, y, o, m)$ consists of the grid location $(x, y)$, vehicle heading $o$ (\fref{fig:steering}(a)), and action $m$ (L: left turn, D: straight, R: right turn).
The edges are defined according to the three actions taken after the current state, as illustrated in \fref{fig:steering}(b).
To generate an efficient path that balances the traversal distance, the friction adequacy on the slopes, surface roughness, and the steering effort, except for the cost of the path length $C_d$, we also define the final A* cost as a weighted summation include the friction cost $C_f$, slope cost $C_\Delta$, roughness cost $C_r$, and steering cost $C_s$ as below.

\myparagraph{Friction Cost} encourages the planner to select routes with higher overall friction and lower viscous resistance.
\begin{equation}
    C_f = \frac{1}{\mu_{\text{PQ}}} + \mu_v^{\text{PQ}},
\end{equation}
where $\mu_{\text{PQ}}$ is the averaged friction coefficient computed using $\mathbf{SL}_\text{P}$ and $\mathbf{SL}_\text{Q}$,
$\mu_v^{\text{PQ}}$ is the averaged viscous coefficient.

\begin{figure}[t]
    \centering
    \includegraphics[width=\linewidth]{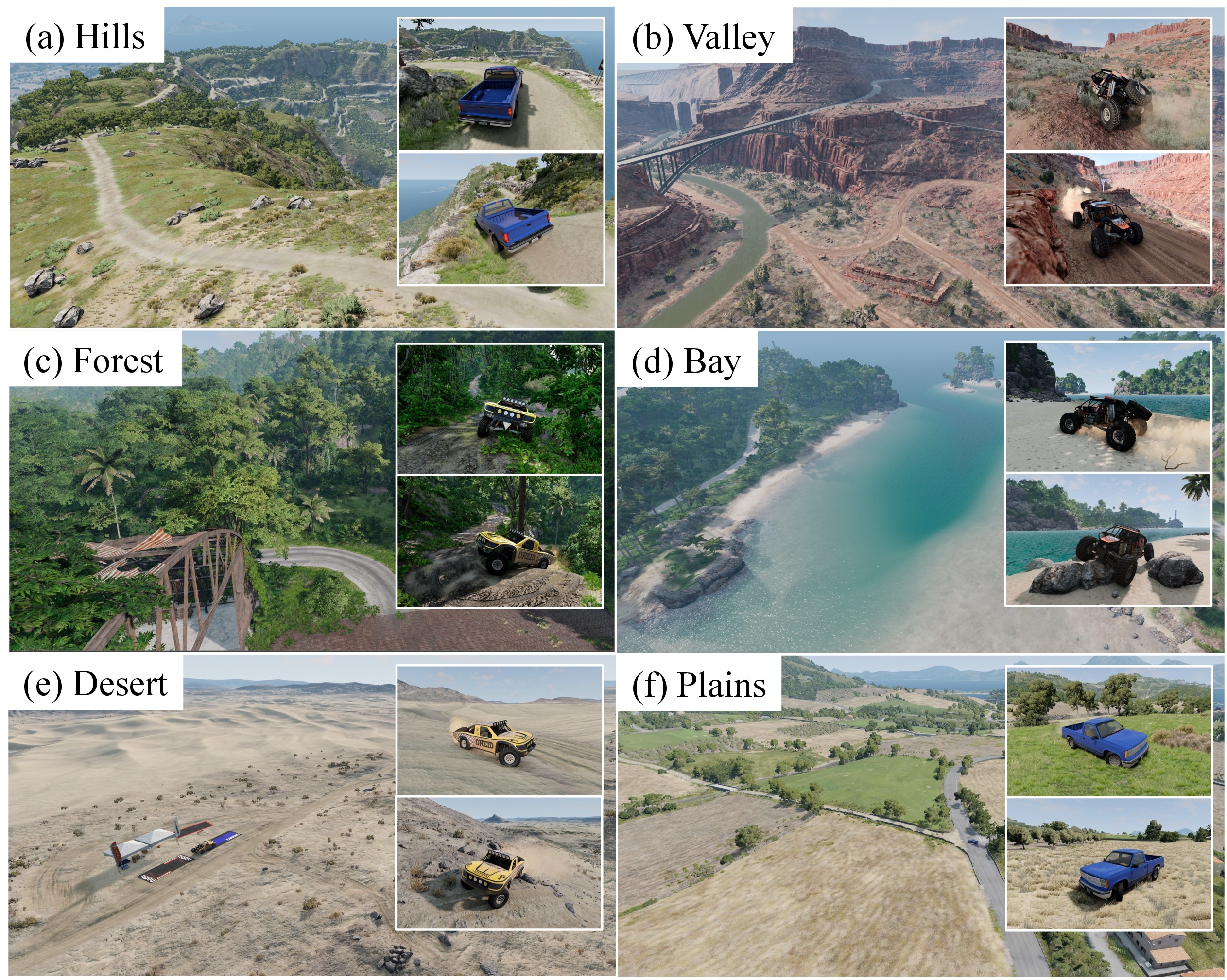}
    \caption{Driving data collection in simulated environments, each featuring diverse terrain types and friction properties.}
    \label{fig:sim_env}
\end{figure}

\myparagraph{Slope Cost} evaluates the vehicle’s ability to maintain control while traversing a slope, affected by friction and slope rate.
\begin{equation}
    C_\Delta = \exp\left(\frac{\delta_{\text{PQ}}}{\mu_{\text{PQ}}}\right) + \exp\left(\frac{\delta_{\perp}}{\mu_{\text{PQ}}} \right),
\end{equation}
where $\delta_{\text{PQ}}$ and $\delta_{\perp}$ are slope rates along and perpendicular to line PQ, respectively, computed from the elevation layer $\mathbf{EL}$.

\myparagraph{Roughness Cost} penalizes traversal through high-roughness regions, which are bumpy and potentially hazardous. $\sigma_{\text{th}}$ is the obstacle threshold, above which the vehicle must avoid.
\begin{equation}
    C_r = 
\begin{cases} 
    (\mathbf{RL}_\text{P}+\mathbf{RL}_\text{Q})/2 & \text{if } \mathbf{RL}_\text{P} < \sigma_{\text{th}} \text{ and } \mathbf{RL}_\text{Q} < \sigma_{\text{th}}, \\ 
    \infty & \text{otherwise.}
\end{cases}
\end{equation}

\myparagraph{Steering Cost} encourages straighter paths as excessive turning reduces the vehicle's speed. The cost depends on the friction coefficient and the turning curvature factor $\eta$,
\begin{equation}
\label{eq:steering_cost}
    C_s = 
\begin{cases} 
    \frac{\eta}{\mu_{\text{PQ}}} & \text{if } o_\text{P} \neq o_\text{Q}, \\ 
    0 & \text{otherwise,}
\end{cases}
\end{equation}
where $\eta=0$, $1$, or $4$ denoting the discretized turning curvature as illustrated in \fref{fig:steering}(c). Specifically, $\eta = 4$ corresponds to consecutive turns in the same direction, $\eta = 1$ to a single turning action, and $\eta = 0$ to opposite turns that cancel after smoothing. In experiments, we found this simple configuration yields effective results without requiring extensive fine-tuning.

For real-time mapping in new environments, the goal position may initially lie outside the mapped area. To handle this, we assign a constant cost $C_\text{unknown}$ per unit distance to unobserved regions to generate potential routes for exploration. The system continuously replans based on the latest map until the vehicle reaches the goal.

\begin{figure}[t]
    \centering
    \includegraphics[width=\linewidth]{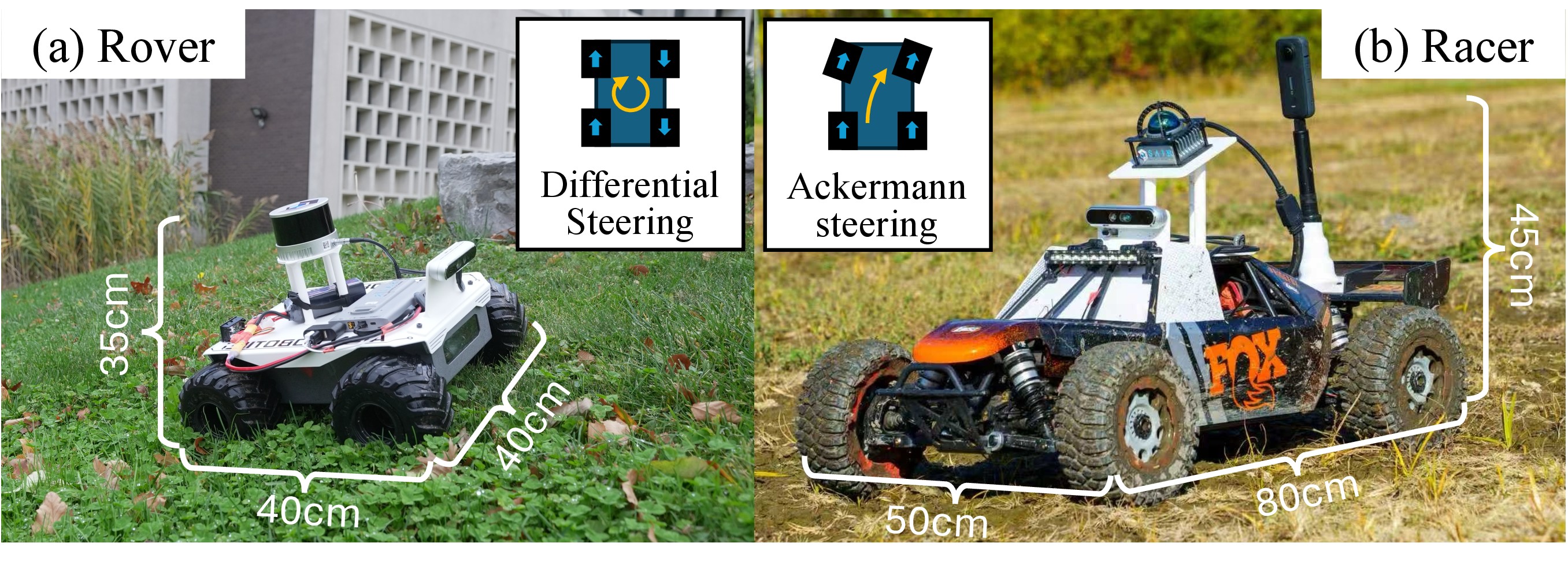}
    \caption{The two real-world robots, \textit{Rover} and \textit{Racer}, are both equipped with LiDAR, camera, IMU, and wheel encoders, but differ in size and steering mechanism. The \textit{Rover} uses differential steering, whereas \textit{Racer} has Ackermann steering.}
    \label{fig:real_env}
\end{figure}

\subsection{Speed Planner} \label{sec:speed_planner}

While the planning module yields a feasible route $\{\mathbf{cp}_1, \mathbf{cp}_2, \cdots, \mathbf{cp}_{N_p}\}$, an effective execution still requires a velocity strategy that is feasible for both terrain conditions and vehicle dynamics. To address this, we introduce a speed planner that generates a physically consistent velocity profile $\{v_1, v_2, \cdots, v_{N_p}\}$ along the planned path by minimizing the total traversal time $T$,
\begin{equation}
    \arg\min_{\{v_i\}} \;\;\; T = \sum_{i=1}^{N_p-1} \frac{2||\mathbf{cp}_{i+1}-\mathbf{cp}_i||}{v_{i+1}+v_i},
\end{equation}
which is subject to three physical constraints conditioned on the available friction, engine capacity, and surface roughness. Therefore, we designed three physical constraints as follows.

\myparagraph{Traction Force Constraint} ensures the composition of forces required for acceleration $F_a$, slope climbing $F_s$, and steering $F_c$ does not exceed the friction $F_f$ the terrain can provide at all checkpoints $i=1\cdots N_p$ on the path:
\begin{equation}\label{eq:froce_cons3}
    \forall i, \;\; F_f^i \geq \sqrt{(F_c^i)^2 + (F_s^i + F_a^i)^2},
\end{equation}
which can be easily computed following Newton's law:
\begin{subequations}\label{eq:force_constraints}
\begin{align}
    F_f^i &= \mu_i M g \cos\theta_i, \label{eq:force_cons1} \\
    F_c^i &= M \cdot \frac{v_i^2}{r_i}, \\
    F_s^i &= M g \sin\theta_i, \\
    F_a^i &= M \cdot \frac{(v_{i+1} - v_i)(v_{i+1} + v_i)}{2\|\mathbf{cp}_{i+1} - \mathbf{cp}_i\|}. \label{eq:force_cons2}
\end{align}
\end{subequations}
where the friction coefficient $\mu_i$, angle of slope $\theta_i$, and terrain roughness $\sigma_i$ are sampled from the terrain property map near the location of checkpoint $\mathbf{cp}_i$. The radius of curvature $r_i$ is computed from the smoothed path at $\mathbf{cp}_i$.

\myparagraph{Engine Saturation Constraint} restricts the longitudinal driving force within the engine's maximum output force $F_{\text{th}}$.
\begin{equation}\label{eq:accel_cons}
    \forall i, \;\; F_s^i + F_a^i \leq F_{\text{th}}.
\end{equation}

\myparagraph{Roughness Constraint} imposes a velocity bound determined by the terrain roughness and wheel radius $r_w$ to maintain stability and traction when traversing bumpy regions.
\begin{equation}
    \forall i, \;\; v_i \leq \frac{(r_w/2)^2}{\sigma_i}.
\end{equation}

We employ the Sequential Least Squares Quadratic Programming (SLSQP) algorithm \citep{kraft1988software} to solve this constrained optimization problem.
A PID controller is employed to generate throttle, brake, and steering commands for the vehicle to follow the planned path and speed profile.

\begin{table*}[!t]\centering
\caption{Comparison of ATE, RRE, and RTE on our simulation dataset. Trajectories are categorized by their traversed terrain types.}\label{tab:sim}
\resizebox{\linewidth}{!}{
\begin{tabular}{c|ccc|ccc|ccc|ccc|ccc}\toprule
\multirow{2}{*}{} &\multicolumn{3}{c|}{TartanDrive} &\multicolumn{3}{c|}{PhysORD} &\multicolumn{3}{c|}{UKF \citep{vosahlik2021self}} &\multicolumn{3}{c|}{PF \citep{vosahlik2021self}} &\multicolumn{3}{c}{\textbf{AnyNav (Ours)}} \\
&ATE &RRE &RTE &ATE &RRE &RTE &ATE &RRE &RTE &ATE &RRE &RTE &ATE &RRE &RTE \\\midrule
Asphalt &1.566 &27.248 &2.524 &1.318 &28.334 &2.594 &2.230 &27.405 &4.479 &2.198 &17.191 &4.483 &\textbf{0.725} &\textbf{9.700} &\textbf{1.523} \\
Rock &0.988 &35.496 &1.380 &1.444 &32.495 &3.026 &2.108 &42.280 &4.340 &1.939 &24.183 &3.875 &\textbf{0.453} &\textbf{11.310} &\textbf{0.943} \\
Dirt &1.056 &30.917 &1.536 &1.037 &34.120 &2.215 &1.992 &28.958 &4.193 &1.712 &29.041 &3.611 &\textbf{0.685} &\textbf{8.116} &\textbf{1.392} \\
Grass &2.560 &33.450 &3.841 &0.890 &39.178 &1.945 &2.387 &33.816 &4.825 &2.162 &26.090 &4.333 &\textbf{0.523} &\textbf{9.153} &\textbf{1.062} \\
Mud &1.316 &29.507 &2.038 &1.290 &35.444 &2.709 &1.631 &34.449 &3.467 &1.387 &21.405 &2.885 &\textbf{0.664} &\textbf{19.480} &\textbf{1.461} \\
Sand &1.092 &26.093 &1.679 &1.161 &25.111 &2.339 &1.728 &18.600 &3.484 &1.554 &20.735 &3.095 &\textbf{0.630} &\textbf{11.032} &\textbf{1.283} \\
Ice &1.652 &29.681 &2.566 &1.068 &24.882 &2.213 &1.028 &12.497 &2.060 &0.679 &15.841 &1.269 &\textbf{0.322} &\textbf{7.567} &\textbf{0.635} \\
\textbf{Overall} &1.462 &30.342 &2.223 &1.173 &31.366 &2.434 &1.872 &28.286 &3.836 &1.662 &22.069 &3.364 &\textbf{0.572} &\textbf{10.908} &\textbf{1.186} \\
\bottomrule
\end{tabular}
}
\end{table*}

\begin{table*}[!t]\centering
\caption{Errors of trajectory estimation on real-world data, including TartanDrive \citep{triest2022tartandrive} and ours Rover and Racer data.}\label{tab:realworld}
\resizebox{0.7\linewidth}{!}{
\begin{tabular}{c|ccc|ccc|ccc}\toprule
\multirow{2}{*}{} &\multicolumn{3}{c|}{TartanDrive Dataset} &\multicolumn{3}{c|}{Rover Dataset} &\multicolumn{3}{c}{Racer Dataset} \\
&ATE &RRE &RTE &ATE &RRE &RTE &ATE &RRE &RTE \\\midrule
TartanDrive &0.754 &5.234 &0.771 &0.187 &38.718 &0.368 &0.784 &39.246 &1.656 \\
PhysORD &0.674 &5.119 &0.586 &0.179 &19.714 &0.354 &0.684 &20.743 &1.453 \\
UKF &0.877 &4.986 &1.800 &0.227 &29.738 &0.458 &0.889 &33.632 &1.699 \\
PF &0.649 &7.320 &1.232 &0.167 &26.005 &0.332 &0.621 &29.721 &1.532 \\
\textbf{AnyNav (Ours)} &\textbf{0.310} &\textbf{2.349} &\textbf{0.543} &\textbf{0.107} &\textbf{5.545} &\textbf{0.177} &\textbf{0.488} &\textbf{11.721} &\textbf{0.901} \\
\bottomrule
\end{tabular}
}
\end{table*}

\section{Experiments}

We next evaluate the effectiveness of the proposed neuro-symbolic friction learning framework in \sref{sec:friction_eval}, the performance of physics-informed navigation in \sref{sec:planning_eval}, and the deployment in unknown environments in \sref{sec:nav_in_unknown}.

\subsection{Experimental Setup}

\subsubsection{Simulated Environments}

We employ BeamNG.tech \citep{beamng_tech} as the driving simulator due to its realistic vehicle-terrain interaction modeling and diverse environments. As shown in \fref{fig:sim_env}, we collect driving data in six environments, each with distinct friction characteristics that affect vehicle dynamics.
To enhance data diversity, we utilize three vehicles (\textit{pickup}, \textit{race truck}, and \textit{rock climber}) driven in autopilot and manual modes. The autopilot mode uses BeamNG’s AI driver with smooth control, whereas the manual mode exhibits aggressive maneuvers such as drifting and full-throttle acceleration. Such maneuvers generate wheel slip with diverse magnitudes and directions, providing richer supervision for learning the friction parameters by exposing the vehicle to a wider range of terrain-vehicle interaction conditions.
We collected a total of 18.9 hours of driving data, including synchronized camera images, LiDAR point clouds, vehicle poses, IMU, and wheel-speed measurements at 10 Hz. The data was split into 80\% for training, 10\% for testing, and 10\% for validation.

\subsubsection{Real-World Robots}

We built two robotic platforms, namely \textit{Rover} and \textit{Racer}, as illustrated in \fref{fig:real_env}, to conduct real-world experiments. The \textit{Rover} platform is based on an Aion R1 chassis. It features direct motor-driven wheels and differential steering, allowing it to turn by varying the relative speeds of the left and right wheels. The \textit{Racer} is adapted from a Losi 1:5 RC truck. It is equipped with compressible suspensions, a high-speed motor, and an Ackermann steering mechanism.
We collected 1.2 and 4.1 hours of driving data on \textit{Rover} and \textit{Racer}, respectively, including synchronized visual, LiDAR, inertial, and wheel speed measurements at 10 Hz. The vehicle poses are tracked using a robust LiDAR-inertial SLAM \citep{zhao2025resilient}.
To facilitate comparison with prior work, we additionally evaluate on the open-source TartanDrive dataset \citep{triest2022tartandrive}, which was collected using a Yamaha all-terrain vehicle.
The differences in their vehicle size and steering mechanisms (differential vs. Ackermann) pose significant challenges to model generalization and sim-to-real transfer.

\begin{figure*}[t]
    \centering
    \begin{subfigure}[b]{0.245\textwidth}
        \includegraphics[width=\textwidth]{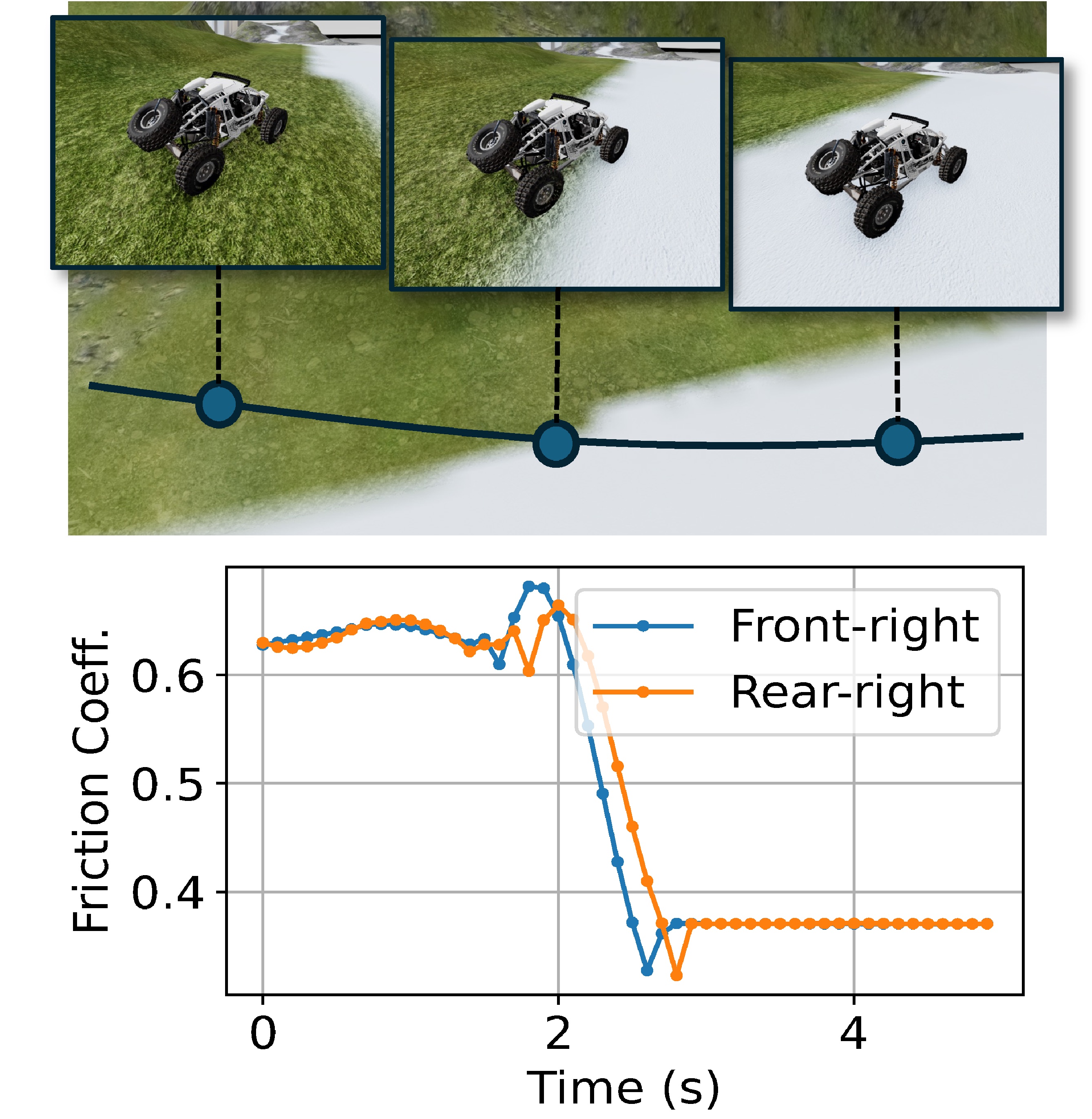}
        \caption{Drive from grass to ice.}
    \end{subfigure}
    \hfill
    \begin{subfigure}[b]{0.245\textwidth}
        \includegraphics[width=\textwidth]{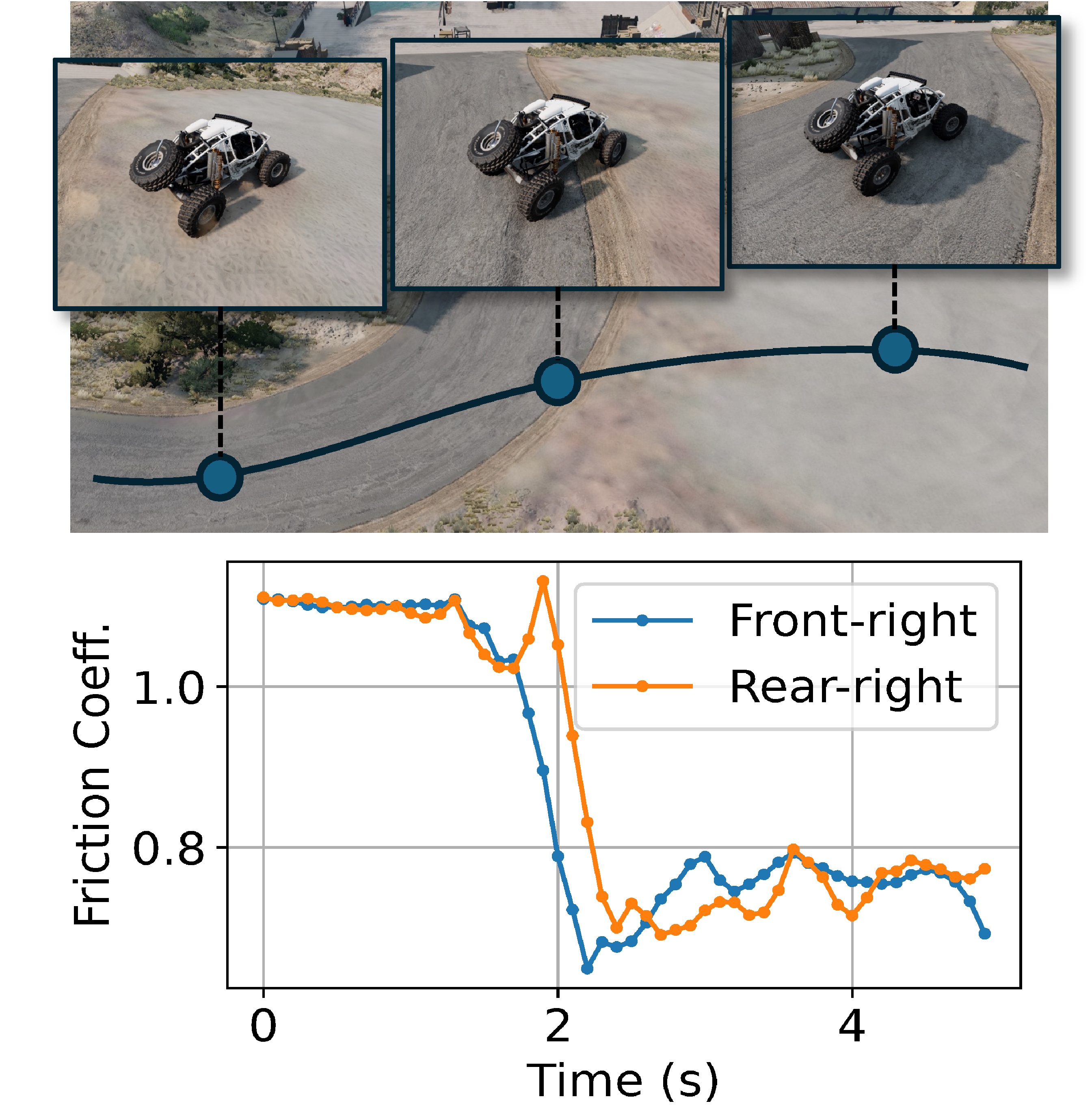}
        \caption{Drive from asphalt to sand.}
    \end{subfigure}
    \hfill
    \begin{subfigure}[b]{0.245\textwidth}
        \includegraphics[width=\textwidth]{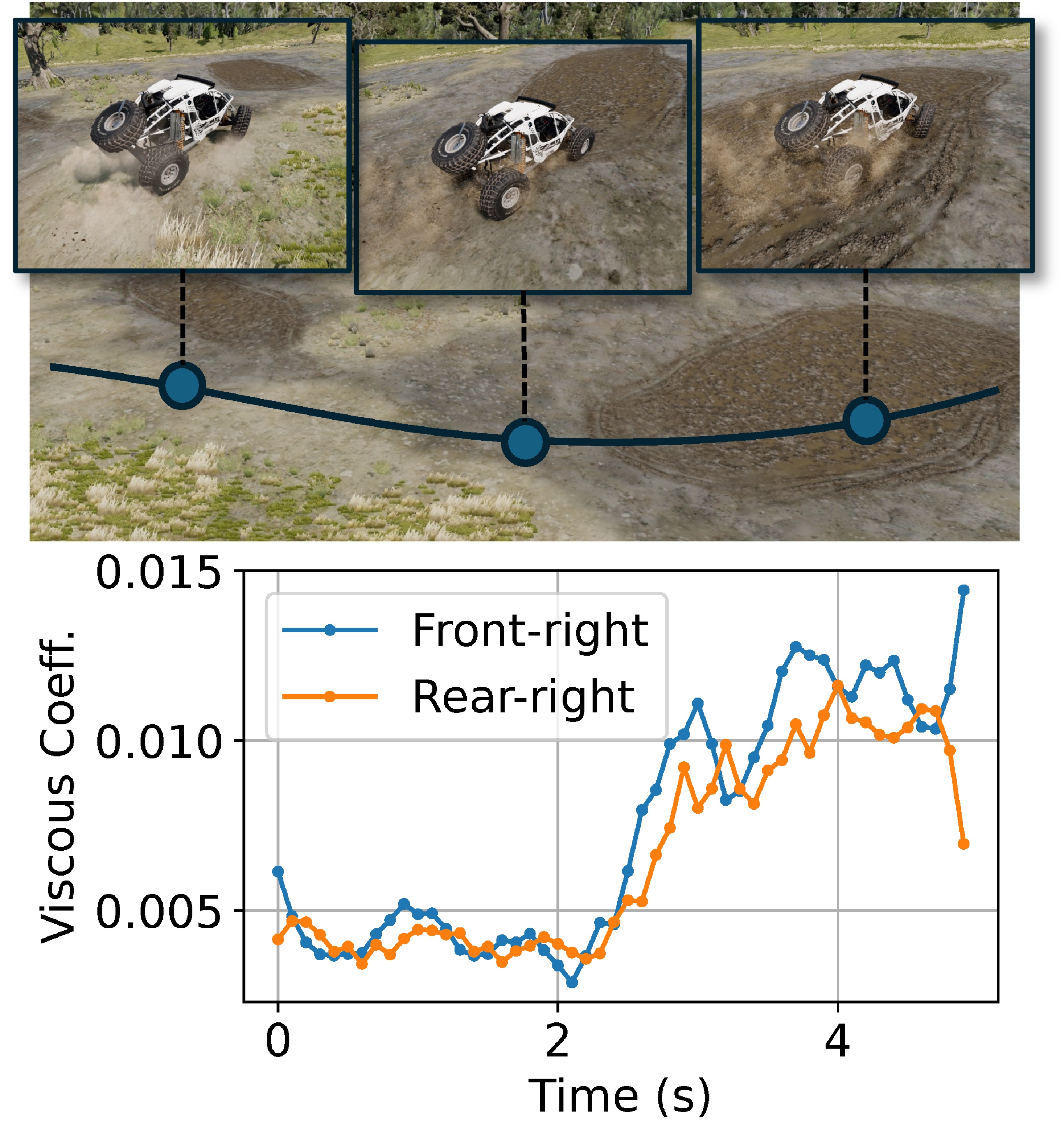}
        \caption{Drive from dirt to mud.}
    \end{subfigure}
    \hfill
    \begin{subfigure}[b]{0.245\textwidth}
        \includegraphics[width=\textwidth]{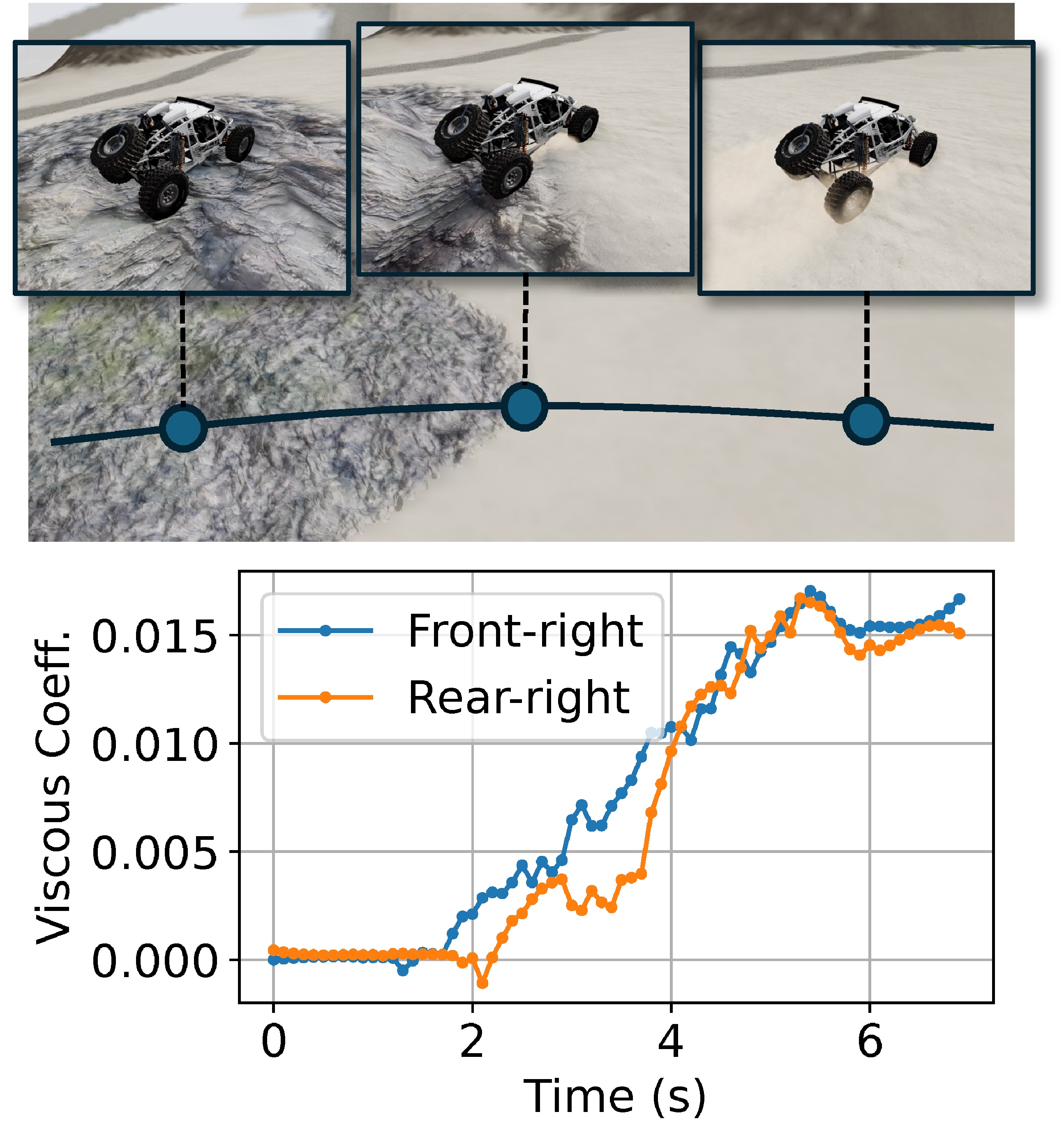}
        \caption{Drive from rock to sand.}
    \end{subfigure}
    \caption{Variation of friction coefficients (a, b) and viscous coefficients (c, d) when driving from one terrain type to another.}
    \label{fig:trans}
\end{figure*}

\begin{figure*}[t]
    \centering
    \includegraphics[width=\linewidth]{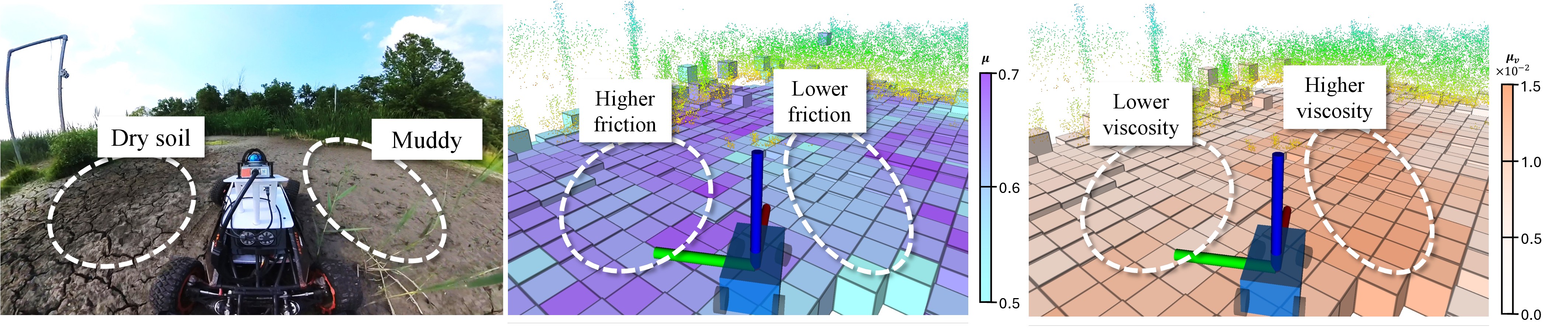}
    \caption{Friction and viscosity coefficients estimated by our model. The friction coefficient is evaluated using the Stribeck model at $v_{\text{rel}}=1\text{m/s}$. The muddy area exhibits lower friction and higher viscosity compared to the dry soil region.}
    \label{fig:real_m1}
\end{figure*}

\begin{figure}[t]
    \centering
    \includegraphics[width=\linewidth]{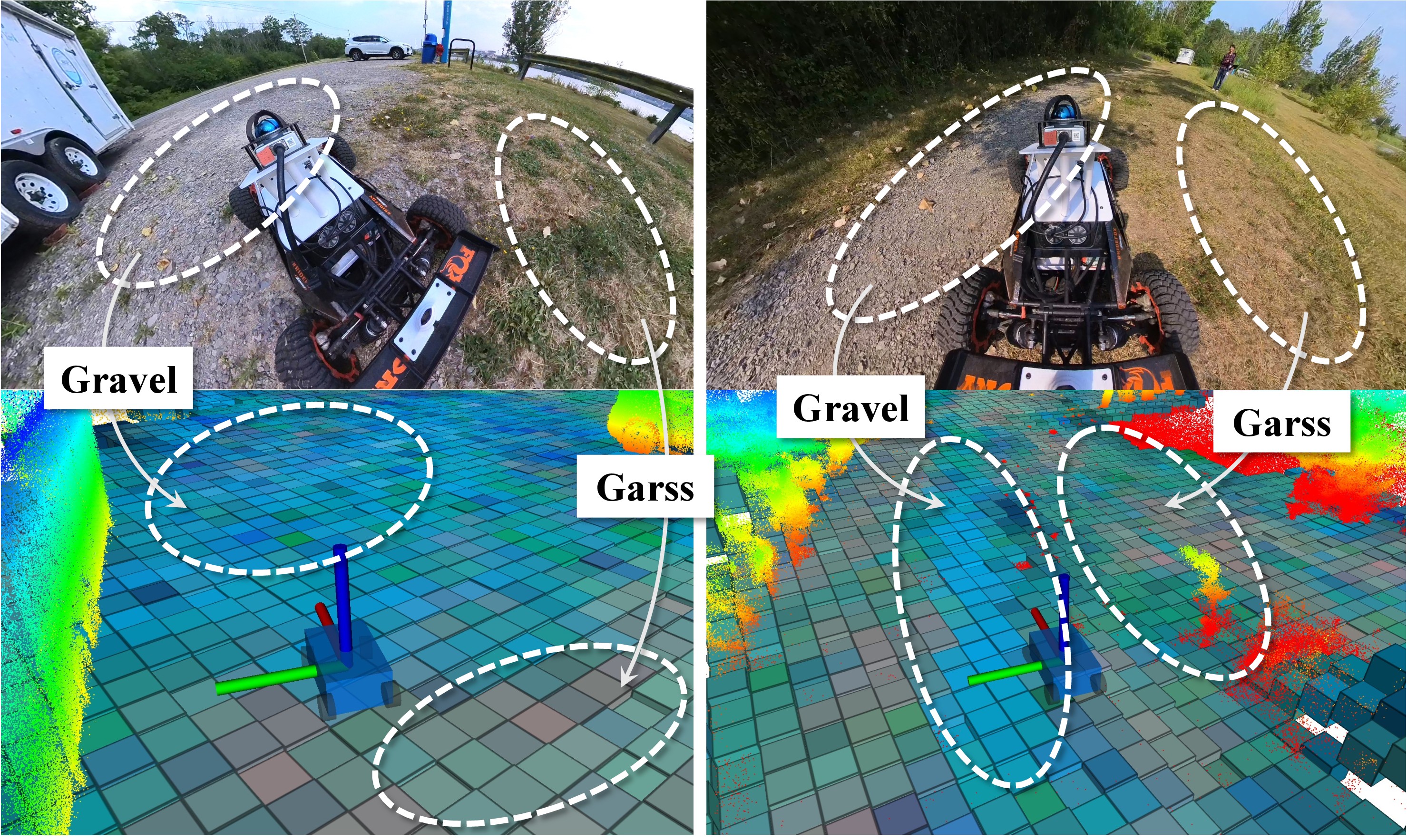}
    \caption{Constructed Stribeck maps visualized by mapping $\mu_v$, $\mu_d$, and $\mu_s$ to the RGB channels, respectively. The gravel and grass regions exhibit clearly different frictional properties.}
    \label{fig:real_m2}
\end{figure}

\subsection{Friction Learning Evaluation} \label{sec:friction_eval}

As the ground-truth friction is not directly measurable, we begin with a qualitative analysis of the learned Stribeck coefficients in \sref{sec:qual}, 
and then perform evaluations on trajectory estimation in \sref{sec:traj} and kinematic prediction in \sref{sec:kine}, as an indirect measure of our friction prediction. The ablation study of our dual-loss and bilevel optimization is presented in \sref{sec:abla}.
Their hyperparameter values are provided in \aref{sec:hyperparam}.

\subsubsection{Qualitative Analysis} \label{sec:qual}

We first examine how the predicted friction coefficient and viscous coefficient vary as individual wheels transition between different terrains. As observed in \fref{fig:trans}, the friction coefficient drops significantly when transitioning from \textit{grass} to \textit{ice} and from \textit{asphalt} to \textit{sand}, while the viscous coefficient increases when transitioning from \textit{dirt} to \textit{mud} and from \textit{rock} to \textit{sand}. These predictions faithfully follow their setups in simulation.

In real-world experiments, \fref{fig:real_m1} illustrates the friction prediction results over a wetland. 
Muddy regions exhibit lower friction and higher viscosity compared to adjacent dry soil, reflecting the expected physical characteristics.
\fref{fig:real_m2} further visualizes the reconstructed Stribeck layer on a hiking trail, where the RGB channels encode the Stribeck parameters $\mu_v$, $\mu_d$, and $\mu_s$, respectively. The resulting color variations highlight the differing frictional properties between gravel-covered and grassy areas.

\subsubsection{Trajectory Estimation} \label{sec:traj}

We next evaluate the precision of the estimated trajectory, as an indirect evaluation of predicted friction. Following \cite{zhao2024physord}'s work, we set time steps $N=20$ and use Absolute Trajectory Error (ATE), Relative Rotation Error (RRE), and Relative Translation Error (RTE) as metrics, evaluated on the testing split of our simulation and real-world data, and the TartanDrive dataset \citep{triest2022tartandrive}.
The results are compared against state-of-the-art learning-based methods, TartanDrive \citep{triest2022tartandrive} and PhysORD \citep{zhao2024physord}, as well as \cite{vosahlik2021self}'s Unscented Kalman Filter (UKF)-based method and its Particle Filter (PF) variant.

\myparagraph{Trajectory Estimation in Simulation}
As shown in \tref{tab:sim}, AnyNav outperforms all baselines across every terrain type in simulation, where trajectories are grouped by their predominant terrain characteristics.
Overall, AnyNav achieves 60.9\% and 51.2\% lower ATE than the learning-based TartanDrive and PhysORD models, and 69.5\% and 65.6\% lower ATE than the filter-based UKF and PF state-estimation methods. 
Since TartanDrive and PhysORD are trained on the same simulation dataset, the performance gains suggest that the improvement primarily arises from the proposed neuro-symbolic learning framework rather than from access to additional training data.

\myparagraph{Trajectory Estimation in Real-world}
In the sim-to-real transfer experiments, \tref{tab:realworld} further confirms that our approach consistently surpasses all baselines.
Compared to PhysORD, which also incorporates physical rules with learning-based methods, our AnyNav achieves 54.0\%, 40.2\%, and 28.7\% lower ATE on the estimated trajectories of the three real-world vehicles, respectively. The performance gain highlights the benefit of explicitly estimating terrain-dependent friction and incorporating it into the vehicle dynamics model, enabling more accurate reasoning about vehicle-terrain interactions.

\begin{figure*}[t]
    \centering
    \begin{subfigure}[b]{1\textwidth}
        \includegraphics[width=\textwidth]{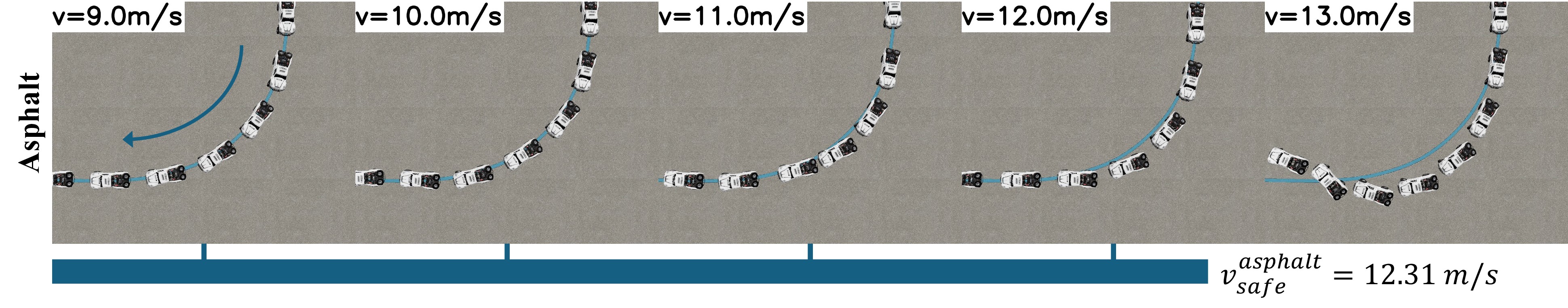}
    \end{subfigure}
    \hfill
    \begin{subfigure}[b]{1\textwidth}
        \includegraphics[width=\textwidth]{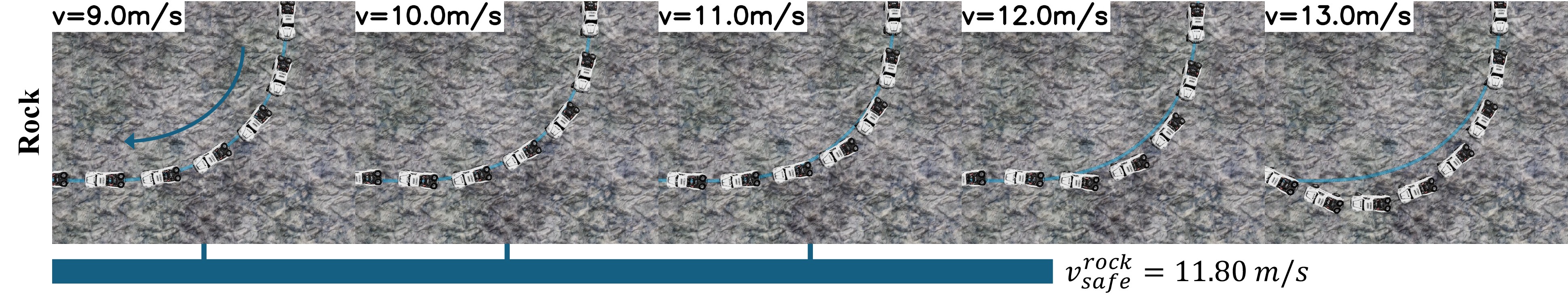}
    \end{subfigure}
    \hfill
    \begin{subfigure}[b]{1\textwidth}
        \includegraphics[width=\textwidth]{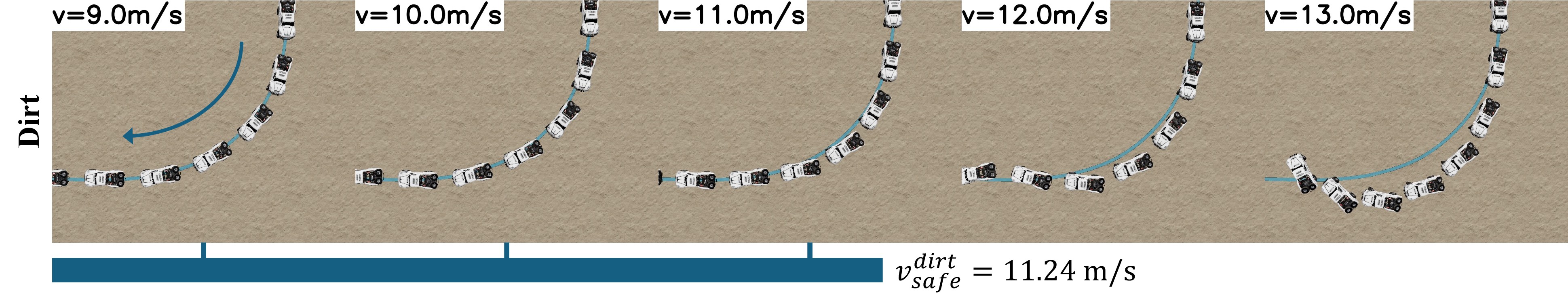}
    \end{subfigure}
    \hfill
    \begin{subfigure}[b]{1\textwidth}
        \includegraphics[width=\textwidth]{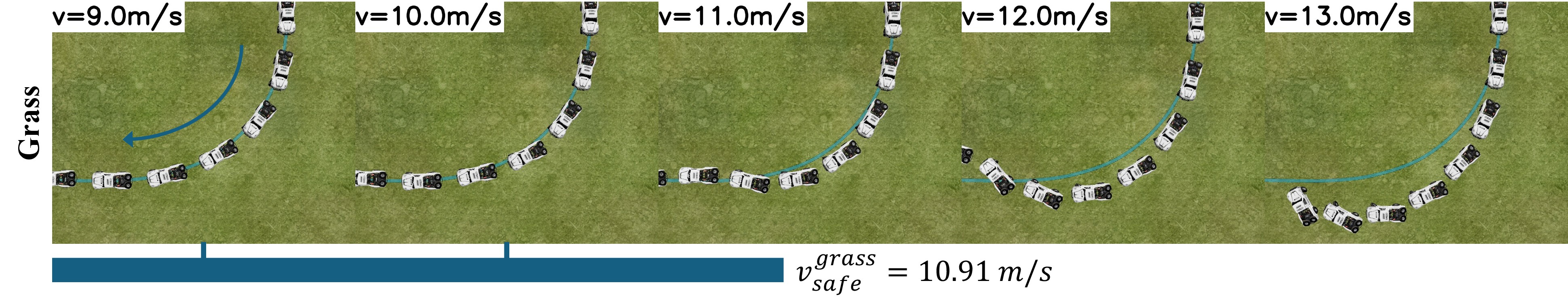}
    \end{subfigure}
    \hfill
    \begin{subfigure}[b]{1\textwidth}
        \includegraphics[width=\textwidth]{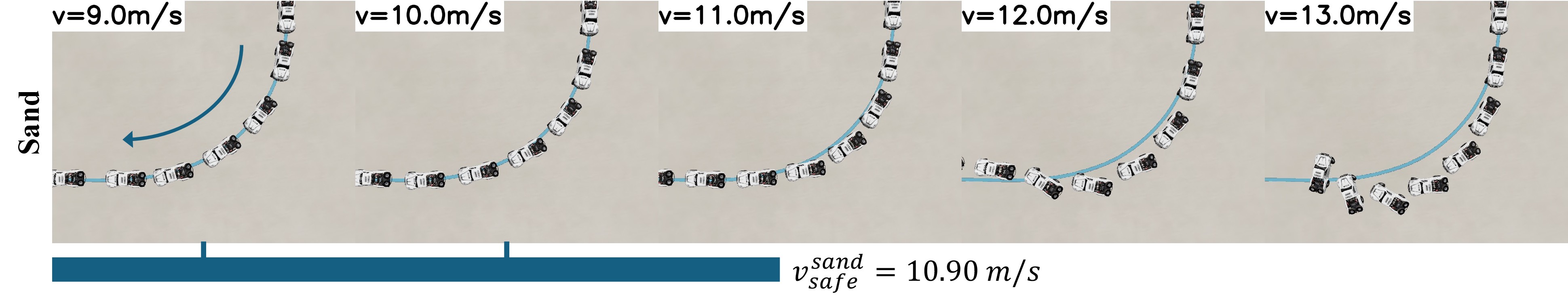}
    \end{subfigure}
    \hfill
    \begin{subfigure}[b]{1\textwidth}
        \includegraphics[width=\textwidth]{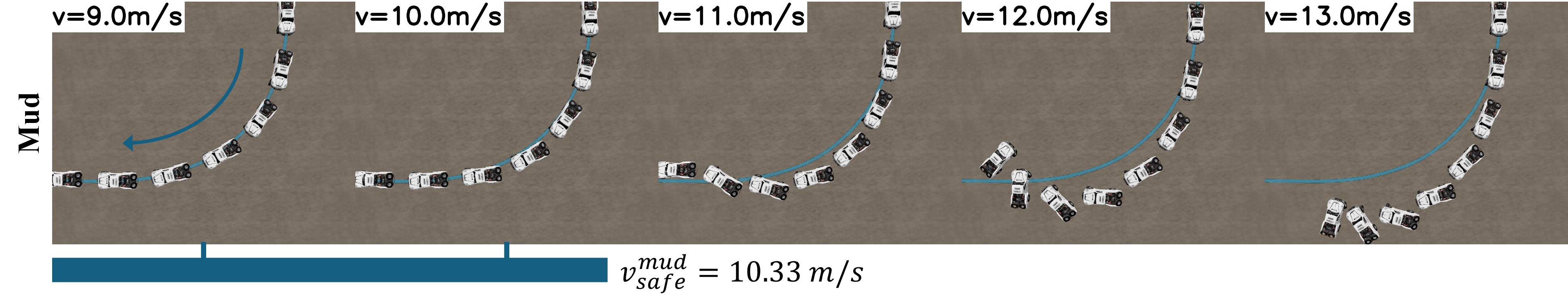}
    \end{subfigure}
    \hfill
    \begin{subfigure}[b]{1\textwidth}
        \includegraphics[width=\textwidth]{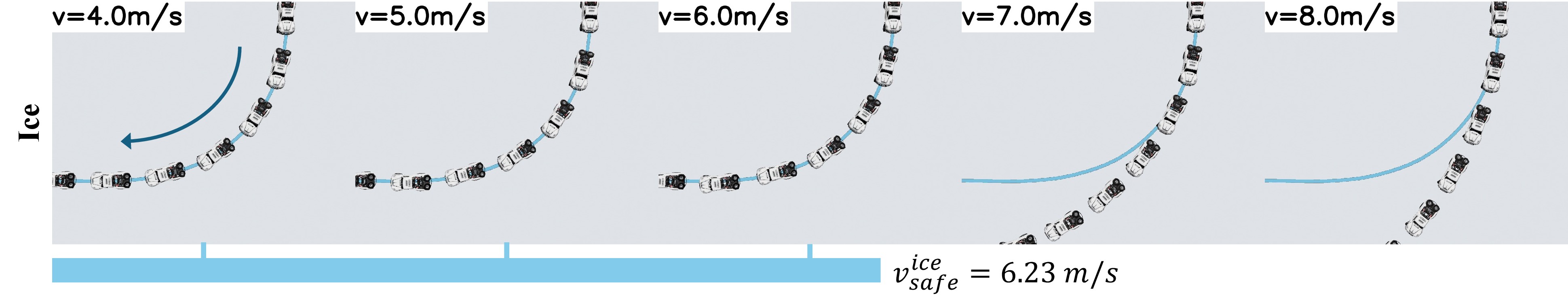}
    \end{subfigure}
    \caption{Top-down views of steering experiments conducted at varying speeds across different terrains. The results show that the predicted maximum safe speed $v_{\text{safe}}$ accurately distinguishes between driftless steering and drifting scenarios.}
    \label{fig:turning}
\end{figure*}

\begin{figure}[!h]
    \centering
    \includegraphics[width=\linewidth]{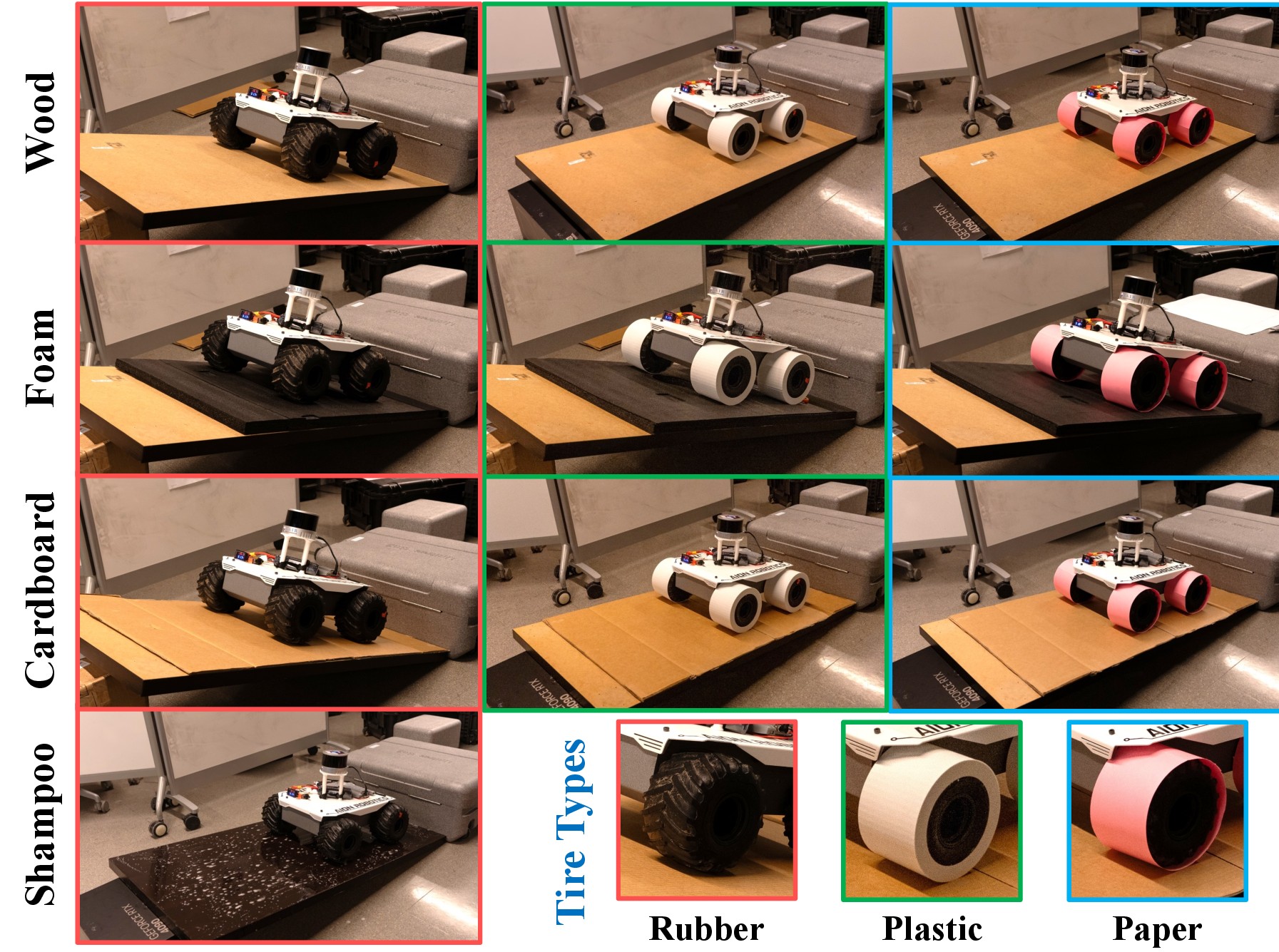}
    \caption{Experimental setup for the slope-climbing test: conducted on four different surface materials using three types of tires, creating a large range of friction conditions.}
    \label{fig:slope}
\end{figure}

\subsubsection{Kinematic Prediction} \label{sec:kine}
We further evaluate the accuracy of friction estimation by examining whether the predicted kinematic behavior aligns with actual motion under different ground conditions. Accordingly, two evaluations are designed: a steering test and a slope-climbing test.

\myparagraph{Steering Test} aims to assess whether the friction estimations correctly reflect the physical limits of cornering stability.
Specifically, we use the predicted Stribeck coefficients $\mathbf{S}_{\text{pred}}^u$ to compute the maximum turning speed $v_{\text{safe}}^u$, at which a vehicle can complete a turn without experiencing significant drift on terrain type $u$. Assuming the available frictional force is fully utilized as the centripetal force, we obtain
\begin{equation}
    M\frac{(v_{\text{safe}}^u)^2}{r_\text{turn}} = \mu^u_{\text{pred}}\cdot Mg \;\; \Rightarrow \;\; v_{\text{safe}}^u = \sqrt{g\cdot r_\text{turn}\cdot \mu_{\text{pred}}^u},
\end{equation}
where $\mu^u_{\text{pred}} = \text{Stribeck}(v_{\text{rel}}; \mathbf{S}_{\text{pred}}^u)$.
We conduct steering trials at various speeds in a simulated environment composed of seven regions. Each region features flat, uniform terrain, ensuring that steering behavior is determined solely by the terrain’s friction. These tests are performed exclusively in simulation, as comparable real-world environments are difficult to construct. We pick a typical slip speed of $v_{\text{rel}}=1\text{m/s}$ and set the turning radius to $r_\text{turn}=20\text{m}$ for all tests.
The results and the predicted safe speeds are visualized in \fref{fig:turning}. It is observed that trial speeds exceeding $v_{\text{safe}}^u$ result in significant drift or even failure to complete the turn, whereas trial speeds below $v_{\text{safe}}^u$ allow the turn to be successfully completed. This outcome highlights both the accuracy of the predicted friction coefficients and their clear physical relevance in modeling terrain-specific dynamics.

\myparagraph{Slope Climbing Test} assesses whether the estimated friction coefficients can reliably predict the maximum slope angle $\theta_{\text{max}}^u$ that a vehicle can ascend. This threshold angle is
\begin{equation}
    \mu^u_{\text{pred}}\cdot Mg\cos\theta_{\text{max}}^u = Mg\sin\theta_{\text{max}}^u \;\; \Rightarrow \;\; \theta_{\text{max}}^u = \arctan(\mu^u_{\text{pred}}),
\end{equation}
where $\mu^u_{\text{pred}}$ is the estimated friction coefficient via the lower-level optimization introduced in \sref{sec:finetuning_real}. The optimization is performed on a sequence of 3-5 minutes of driving data on ground type $u$.
We conducted real-world tests on slopes with different surface materials, including \textit{wood, foam, cardboard}, and a \textit{shampoo-covered board}. To further assess generalizability, we replaced the standard rubber tire surface with alternative materials such as plastic (3D printed) and paper. The experimental setups are shown in \fref{fig:slope}. As summarized in \tref{tab:slope}, the predicted maximum slope angles closely match the observed results.
Notably, the predictions are more accurate for low-friction surfaces. This is because the vehicle exhibits more pronounced slipping when friction is low, providing a greater number of valid data points for learning the Stribeck curve.

\begin{table}[t]\centering
\caption{The predicted and measured maximum climb angles (in degrees) across various ground and wheel materials.}\label{tab:slope}
\resizebox{\linewidth}{!}{
\begin{tabular}{c|cc|cc|cc}\toprule
\multirow{2}{*}{} &\multicolumn{2}{c|}{Rubber} &\multicolumn{2}{c|}{Plastic} &\multicolumn{2}{c}{Paper} \\
&Pred. &Meas. &Pred. &Meas. &Pred. &Meas. \\\midrule
Wood &39.2 &43 &15.8 &16 &13.4 &14 \\
Foam &41.2 &46 &23.9 &23 &27.9 &25 \\
Cardboard &39.7 &44 &11.1 &10 &11.0 &12 \\
Shampoo &16.4 &15 &- &- &- &- \\
\bottomrule
\end{tabular}
}
\end{table}

\begin{table}[t]\centering
\caption{Ablation study of the contributions of the acceleration loss $L_\text{acc}$ and prior knowledge loss $L_\text{prior}$ for model training.}
 \label{tab:ablation_sim}
\resizebox{0.7\linewidth}{!}{
\begin{tabular}{c|ccc}\toprule
&ATE &RRE &RTE \\\midrule
$L_\text{acc}$ Only &0.685 &15.558 &1.394 \\
$L_\text{prior}$ Only &0.658 &14.839 &1.405 \\
\textbf{AnyNav (Ours)} &\textbf{0.583} &\textbf{11.420} &\textbf{1.216} \\
\bottomrule
\end{tabular}
}
\end{table}

\subsubsection{Ablation Studies} \label{sec:abla}

We conduct ablation studies to evaluate the contribution of each loss term during training and examine the effectiveness of jointly optimizing the vehicle model and sensor noise in sim-to-real transfer.

\myparagraph{Ablation on Loss Terms}
We first analyze the contribution of the training loss terms by comparing performance when using only the acceleration loss, only the prior-knowledge loss, and both losses jointly. The results, presented in \tref{tab:ablation_sim}, show that our method yields 17.5\% and 12.9\% better performance in terms of ATE compared to using the acceleration loss and the prior-knowledge loss individually. This demonstrates the complementary nature of the two losses in improving overall model accuracy.

\myparagraph{Ablation on Bilevel Optimization Variables}
We further investigate the sim-to-real transfer process by isolating the effects of individual variables in bilevel optimization. Specifically, we separately fix the wheel speed to its noisy raw measurements and the vehicle’s inertia matrix to its initial estimate, rather than optimizing them jointly. We also evaluate the performance of the model without any sim-to-real transfer. The results in \tref{tab:ablation_real} show that our method outperforms all these variants, achieving on average 21.8\% lower ATE than the variant with fixed wheel speed and 12.2\% lower ATE than the variant with fixed inertia matrix.
This highlights the effectiveness of jointly optimizing sensor noise parameters and the vehicle model together with friction predictions within the bilevel optimization framework.

\subsection{Navigation Evaluation} \label{sec:planning_eval}

In this section, we evaluate the path feasibility and time efficiency of AnyNav in complex off-road environments. 
To highlight the importance of incorporating physical properties, we assume that complete terrain maps are prebuilt, providing sufficient information for global planning. 
This setting is important for applications like agricultural automation, where the entire farmland can be mapped once and used for all navigation tasks. Real-time planning in unknown environments will be presented in \sref{sec:nav_in_unknown}.

\begin{table*}[t]\centering
\caption{Ablation study comparing no sim-to-real transfer and fixing wheel speed and inertia matrix during the bilevel optimization.}\label{tab:ablation_real}
\resizebox{0.75\linewidth}{!}{
\begin{tabular}{c|ccc|ccc|ccc}\toprule
\multirow{2}{*}{} &\multicolumn{3}{c|}{TartanDrive Dataset} &\multicolumn{3}{c|}{Rover Dataset} &\multicolumn{3}{c}{Racer Dataset} \\
&ATE &RRE &RTE &ATE &RRE &RTE &ATE &RRE &RTE \\\midrule
No sim-to-real transfer &0.675 &4.450 &1.212 &0.117 &7.315 &0.193 &0.645 &23.666 &1.217 \\
Fixed wheel speed &0.656 &3.988 &1.177 &0.115 &5.896 &0.188 &0.517 &12.087 &0.953 \\
Fixed inertia matrix &0.366 &3.478 &0.641 &0.109 &6.793 &0.183 &0.605 &23.472 &1.153 \\
\textbf{AnyNav (Ours)} &\textbf{0.310} &\textbf{2.349} &\textbf{0.543} &\textbf{0.107} &\textbf{5.545} &\textbf{0.177} &\textbf{0.488} &\textbf{11.721} &\textbf{0.901} \\
\bottomrule
\end{tabular}
}
\end{table*}

\begin{table*}[t]\centering
\caption{Comparison of success rate, average time, and time ratio across two Maps. Note that the Average Time and Time Ratio are computed only on successful tasks. The best success rates are marked in bold.}\label{tab:missions}
\resizebox{\linewidth}{!}{
\begin{tabular}{c|c|ccc|ccc}\toprule
\multicolumn{2}{c|}{} &\multicolumn{3}{c|}{Garden} &\multicolumn{3}{c}{Island} \\
\multicolumn{2}{c|}{} &Succ. Rate &Avg. Time &Time Ratio &Succ. Rate &Avg Time &Time Ratio \\\midrule
\multirow{2}{*}{\makecell{Traversibility\\Estimation}} &Binary \citep{ramirez2024real} &72\% &31.661 &1.116 &60\% &46.544 &1.208 \\
&URA* \citep{moore2023ura} &26\% &53.774 &- &8\% &62.605 &- \\\midrule
\multirow{5}{*}{\makecell{Friction\\Sensitivity}} &Uniform friction coeff. $\mu=0.2$ &90\% &42.820 &0.969 &92\% &73.678 &0.962 \\
&Uniform friction coeff. $\mu=0.5$ &82\% &29.765 &1.078 &92\% &46.183 &1.063 \\
&Uniform friction coeff. $\mu=0.8$ &82\% &28.197 &1.127 &82\% &42.408 &1.208 \\
&Scale predicted friction coeff. by 0.5 &\textbf{100\%} &39.260 &0.997 &92\% &55.217 &1.004 \\
&Scale predicted friction coeff. by 1.5 &86\% &29.620 &1.158 &74\% &38.569 &1.181 \\\midrule
\multirow{2}{*}{\makecell{Velocity\\Sensitivity}} &Scale planned velocity by 0.75 &\textbf{100\%} &35.830 &0.942 &\textbf{94\%} &49.620 &0.986 \\
&Scale planned velocity by 1.25 &82\% &27.638 &1.230 &80\% &38.032 &1.286 \\\midrule
\multicolumn{2}{c|}{Ours without steering cost} &96\% &32.304 &1.064 &88\% &44.405 &1.083 \\
\multicolumn{2}{c|}{\textbf{AnyNav (Ours)}} &\textbf{100\%} &29.648 &1.037 &\textbf{94\%} &41.942 &1.099 \\
\bottomrule
\end{tabular}
}
\end{table*}

\subsubsection{Quantitative Evaluation}
\label{sec:quat_known_map}

The evaluation is conducted in \textit{Garden} and \textit{Island}, two challenging environments containing diverse terrain types, varying slopes, and uneven vegetation. Terrain property maps are computed using top-down RGB-D images captured at 50-meter intervals throughout each environment. The friction estimation network is trained beforehand and remains fixed during evaluation. To avoid data leakage and overfitting, neither environment is included in the training set of the network. For each map, 50 start-goal pairs are randomly sampled and manually verified to ensure feasibility, i.e., the start and goal points are not placed on cliffs or other inaccessible regions. All methods are evaluated using the same start-goal pairs for fairness.

We evaluate the navigation results using three metrics. The \textit{success rate} measures the percentage of successfully completed tasks, reflecting the feasibility and safety of the planned paths and speed profiles. The \textit{average time} denotes the mean travel time of completed tasks, indicating navigation efficiency.
The \textit{time ratio} is defined as the actual traversal time measured during execution divided by the predicted traversal time computed from the planned speed profile. A value close to 1 indicates that the planned speed profile is physically achievable and accurately reflects the vehicle's capability. In contrast, a time ratio greater than 1 suggests that the planned speed is overly aggressive relative to the available terrain friction, causing wheel slip, reduced acceleration, drifting, or other effects that increase the actual traversal time.
We evaluated our method against four groups of baselines to provide a comprehensive analysis of its advantages in various aspects, described as follows.

\begin{figure}[!b]
    \centering
    \includegraphics[width=\linewidth]{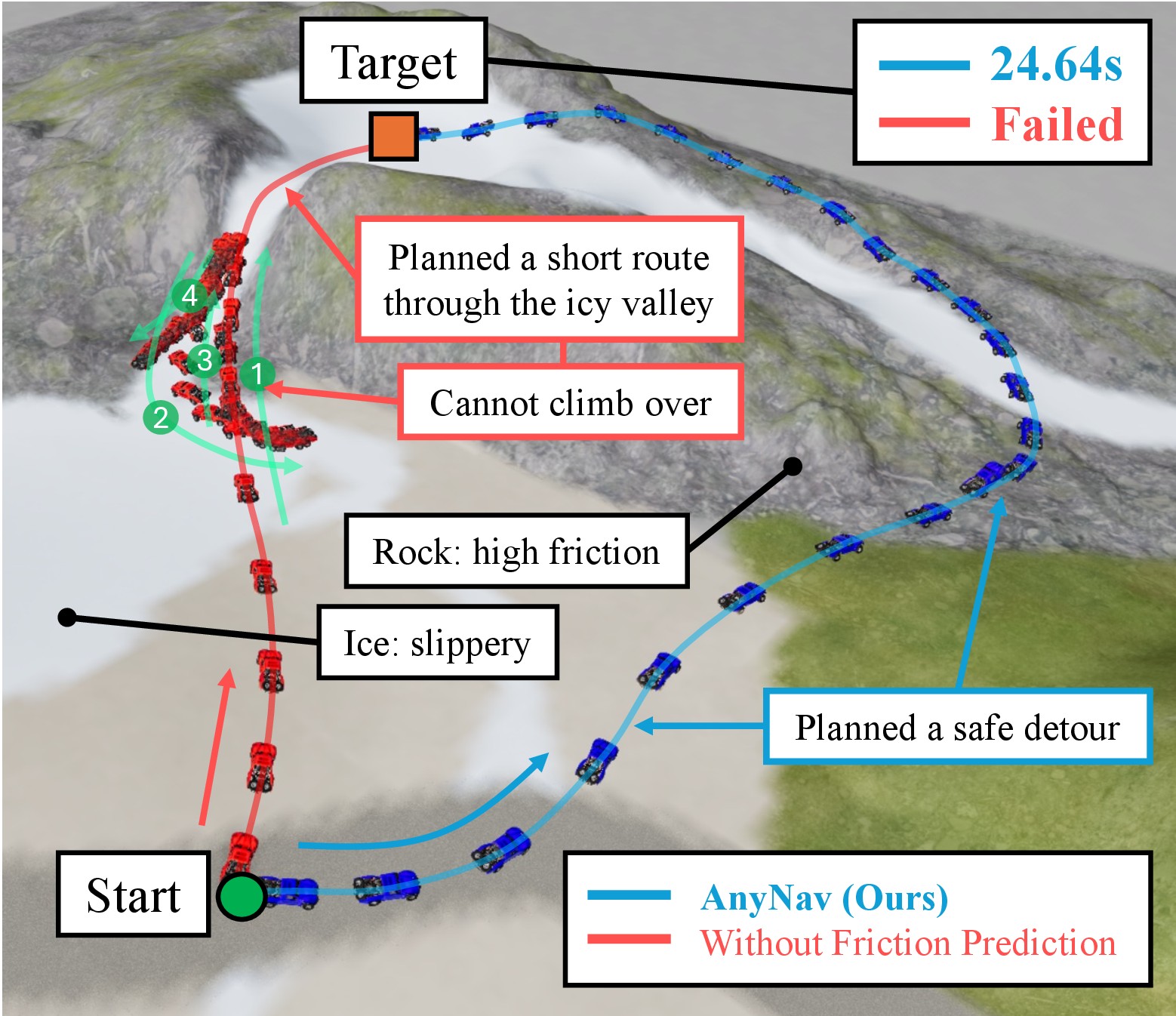}
    \caption{A navigation task where the target is atop an icy crater. The planner without friction prediction (red) attempted to use a shorter path but failed due to the low friction in the icy valley. In contrast, AnyNav (blue) suggests a safe detour with enough friction, successfully completing the task.}
    \label{fig:plan11}
\end{figure}

\myparagraph{Comparison to Traversability Estimation Baselines}
The first group of baselines involves path planning methods based on estimated traversability maps. We first implement an approach similar to \citep{ramirez2024real}, generating a traversability map in which obstacles and icy regions are marked as non-traversable. Paths are then planned using the standard A* algorithm to avoid these areas. This baseline is referred to as ``Binary'' because it treats terrain as either traversable or non-traversable, without modeling continuous variations in friction or roughness properties. Additionally, we implemented URA* \citep{moore2023ura}, which uses a learning-based method to identify traversable and non-traversable regions from BEV images, followed by an A*-like algorithm for route planning.
As shown in \tref{tab:missions}, AnyNav achieves 28\% and 34\% higher success rates than the binary traversability-map baseline in the two environments. URA* performs significantly worse, especially on \textit{Island}, because its learning model fails to generalize to such complex environments. These results highlight the effectiveness of using continuous, physics-informed traversability representations in AnyNav.

\begin{figure*}
    \centering
    \includegraphics[width=\linewidth]{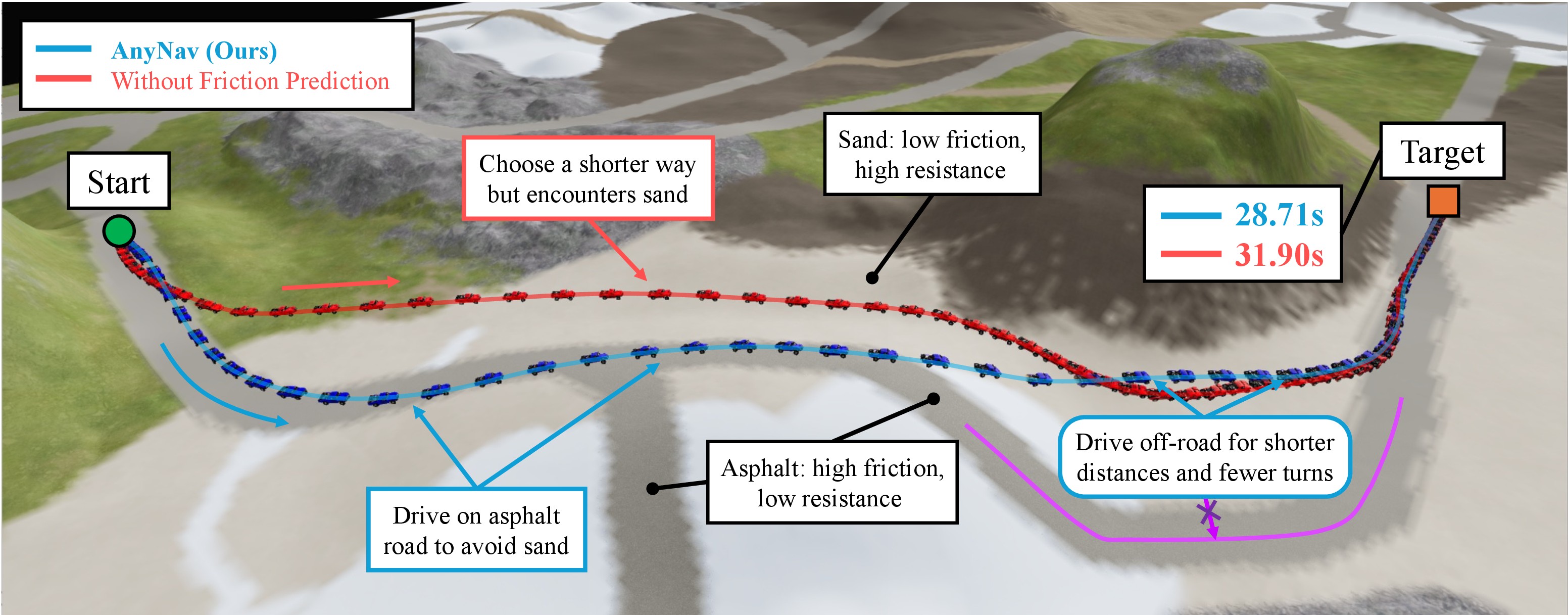}
    \caption{An example highlights the advantage of AnyNav, which optimizes travel time by selecting a low-resistance route.}
    \label{fig:plan12}
\end{figure*}

\begin{figure*}
    \centering
    \includegraphics[width=\linewidth]{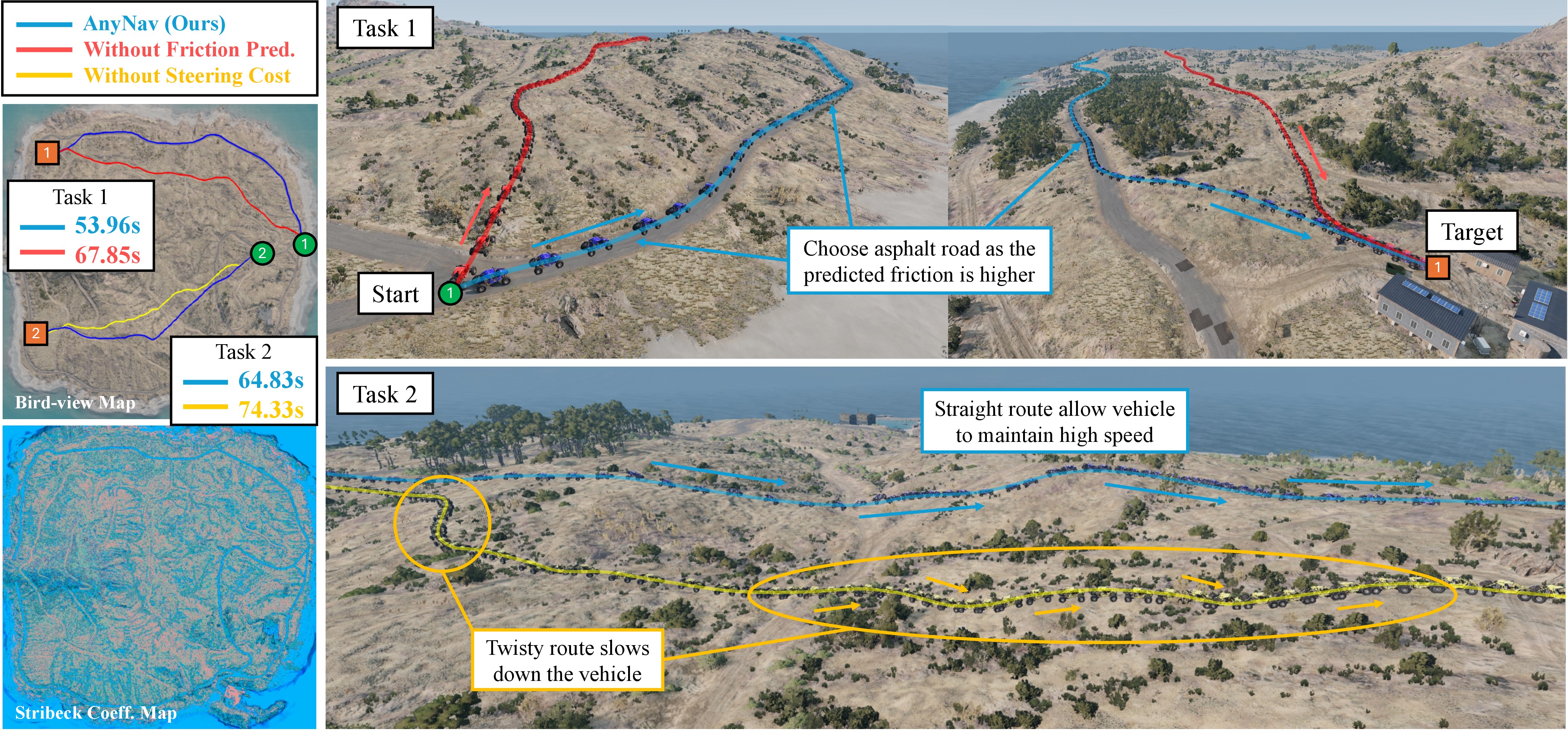}
    \caption{Two navigation examples on the \textit{Island} map. Task 1 highlights how incorporating friction knowledge into planning reduces travel time. Task 2 showcases the efficiency of our steering-minimized strategy in improving average speed.}
    \label{fig:plan2}
\end{figure*}

\myparagraph{Comparison to Navigation with Modified Friction}
To evaluate the importance of incorporating accurate friction estimations for planning, we introduce a second group of baselines that alter the friction information provided to the planners. In the first three baselines, the friction coefficient is set to constant values of 0.2, 0.5, and 0.8 across the entire map, eliminating any terrain-specific information. 
To verify the effects of biased friction predictions, in the latter two baselines, we scale the predicted friction coefficients by 0.5 and 1.5, respectively.
It is observed in \tref{tab:missions} that all such modifications result in decreases in success rates and increases in average travel times. Specifically, when the friction coefficient is set to a low value (e.g., $\mu=0.2$ or $\mu=0.5\mu_{\text{pred}}$), the navigation system becomes overly cautious, resulting in significantly reduced speeds and therefore increased average travel times by 31.7\%-75.7\%. Conversely, when the friction coefficient is set to a high value (e.g., $\mu=0.8$ or $\mu=1.5\mu_{\text{pred}}$), the navigation system becomes overly confident, leading to aggressive plans that cause a significant drop in success rate by 12\%-20\%. These outcomes demonstrate that AnyNav produces friction estimates that faithfully reflect the underlying physical constraints, allowing the path and speed planners to generate safe and efficient navigation strategies.

\begin{figure*}[!t]
    \centering
    \includegraphics[width=\linewidth]{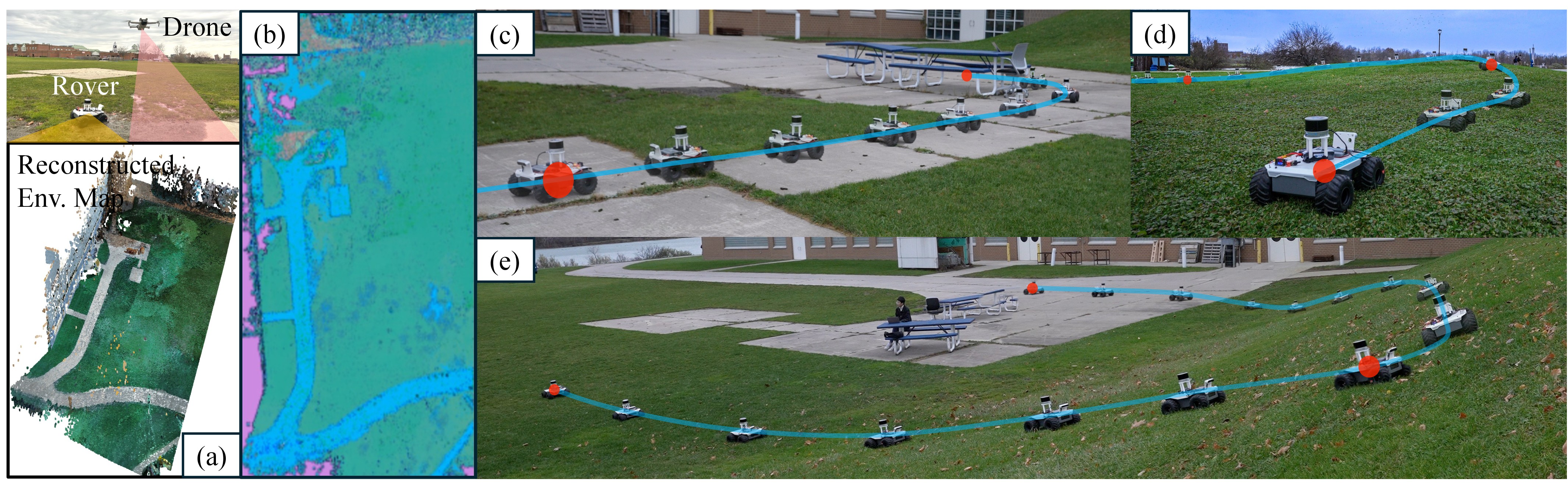}
    \caption{Real-world pre-mapping and navigation tests. (a) The environment map is reconstructed through MVS using drone-captured images. (b) The Stribeck coefficient is estimated using the neuro-symbolic module. Different terrain surfaces, such as lawn and road, are clearly differentiated by their respective friction characteristics. (c-e) Multiple navigation tasks are executed. The red points are selected waypoints, and the blue lines are planned trajectories.}
    \label{fig:realworld}
\end{figure*}

\myparagraph{Comparison to Driving at Scaled Velocities}
To evaluate the robustness and time efficiency of our speed profile, we introduce a third group of baselines that scale the planned speed by 0.75$\times$ and 1.25$\times$, illustrating the impact of driving slower and faster than desired.
In \tref{tab:missions}, compared to the 0.75$\times$ baseline, our planned speed achieves 20.9\% and 18.3\% shorter travel time. Compared to the 1.25$\times$ baseline, our method achieves 18\% and 14\% higher success rates.
Note that the average time is calculated only for successful tasks, making it a meaningful comparison only when success rates are similar. These results demonstrate that our planned speed is nearly optimal: deviating from it leads to a drop in performance. Additionally, the time ratio for the 1.25$\times$ baseline is significantly greater than 1, while that for ours is approximately 1. This indicates that our speed planner effectively leverages the terrain's friction while remaining within the vehicle's operational limits.

\myparagraph{Comparison to Planning Without Steering Cost}
The fourth baseline excludes the steering cost \eqref{eq:steering_cost} from the path planning process. 
Compared to this baseline, AnyNav achieved 9.0\% and 5.9\% shorter travel time in the two environments. This highlights the effectiveness of our steering-minimization strategy in enhancing overall navigation time efficiency.

\begin{table}[t]\centering
\caption{\textbf{Navigation results in the mapping-from-scratch mode.} Each task begins with no environmental knowledge.} \label{tab:mfs}
\resizebox{\linewidth}{!}{ 
\begin{tabular}{c|cc|cc}\toprule
&\multicolumn{2}{c|}{Garden} &\multicolumn{2}{c}{Island} \\
&Succ. Rate &Avg. Time &Succ. Rate &Avg. Time \\\midrule
Binary &60.00\% &76.85 &46.00\% &116.58 \\
MPPI-based &72.00\% &80.33 &54.00\% &136.38 \\
\textbf{AnyNav (Ours)} &\textbf{74.00\%} &85.55 &\textbf{84.67\%} &122.04 \\
\bottomrule
\end{tabular}
}
\end{table}

\subsubsection{Qualitative Comparisons}

We show several representative trajectories planned with and without friction predictions to demonstrate the performance gap between AnyNav and a uniform friction baseline. We set $\mu=0.5$ for the baseline as it performs best among the uniform-friction baselines reported in \tref{tab:missions}.

\begin{table}[t]\centering
\caption{\textbf{Navigation results in the online accumulative mapping mode.} The system starts in unknown environments, incrementally builds a global terrain map while executing the navigation tasks. Subsequent tasks can leverage information from mapped regions explored during earlier tasks.} \label{tab:am}
\resizebox{\linewidth}{!}{ 
\begin{tabular}{c|cc|cc}\toprule
&\multicolumn{2}{c|}{Garden} &\multicolumn{2}{c}{Island} \\
&Succ. Rate &Avg. Time &Succ. Rate &Avg. Time \\\midrule
Binary &66.00\% &78.21 &51.33\% &114.63 \\
MPPI-based &78.00\% &88.36 &56.00\% &145.57 \\
\textbf{AnyNav (Ours)} &\textbf{85.33\%} &86.59 &\textbf{88.67\%} &121.82 \\
\bottomrule
\end{tabular}
}
\end{table}

\myparagraph{Failure Cases Arising from Ignoring Friction}
\fref{fig:plan11} illustrates a navigation task in \textit{Garden}, where the target point is located atop an icy crater. Without considering the terrain-specific friction properties, the baseline selects a shorter route through the icy valley (red line). This choice results in navigation failure, as the icy region is too slippery for the vehicle to ascend (green arrows). In contrast, our physics-informed planner, which reasons about both friction levels and slope angles, selects a detour that avoids most of the icy regions and utilizes high-friction rocky areas to ascend (blue line). This physically feasible route enables the vehicle to reach the target successfully, highlighting the importance of modeling friction for reliable navigation.

\myparagraph{Increased Traversal Time Without Friction Awareness}
\fref{fig:plan12} illustrates another navigation task in the \textit{Garden}. The baseline without friction knowledge selects a shorter route across the sand (red line), while our AnyNav generates a longer path on an asphalt road (blue line). This decision is based on the navigation system's understanding that sand is more slippery and has higher resistance, while the asphalt road allows for a higher average speed. As a result, our planned route is 11.1\% faster by considering physical knowledge.
Similarly, in Task 1 in \fref{fig:plan2}, AnyNav finished 25.7\% faster than the baseline. These examples highlight the advantage of incorporating physics-based friction predictions for optimizing travel time.

\myparagraph{Increased Execution Time Without the Steering Cost}
Task 2 in \fref{fig:plan2} highlights the effectiveness of our steering-minimization design. Without the steering cost, the planned path is more winding (yellow line), which reduces the vehicle's average speed. In contrast, AnyNav generates a smoother path with more straight segments (blue line), which is $14.7\%$ faster. This highlights the effectiveness of our approach in incorporating steering-related costs to significantly improve overall travel efficiency.

\begin{figure*}[t]
    \centering
    \includegraphics[width=\linewidth]{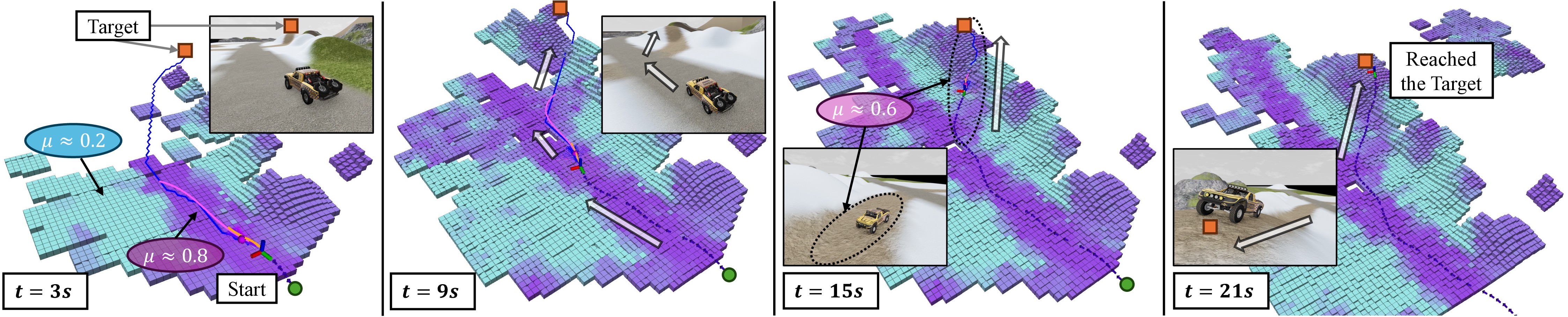}
    \caption{Example of navigating in an initially unknown environment. The friction coefficient map gradually expands as the vehicle moves, while the route is continuously replanned based on the updated map. At $t=3s$, the vehicle moves forward along the road as the sides are too slippery. The path segment outside the mapped region is tentative, providing only a rough direction, and will be updated later as the map expands. At $t=9s$, the vehicle clearly observes the dirt road leading uphill, and the planner selects a path along it. At $t=15s$, the vehicle climbs the hill, and by $t=21s$ it reaches the goal.}
    \label{fig:sim_fm}
\end{figure*}

\begin{figure*}[t]
    \centering
    \includegraphics[width=\linewidth]{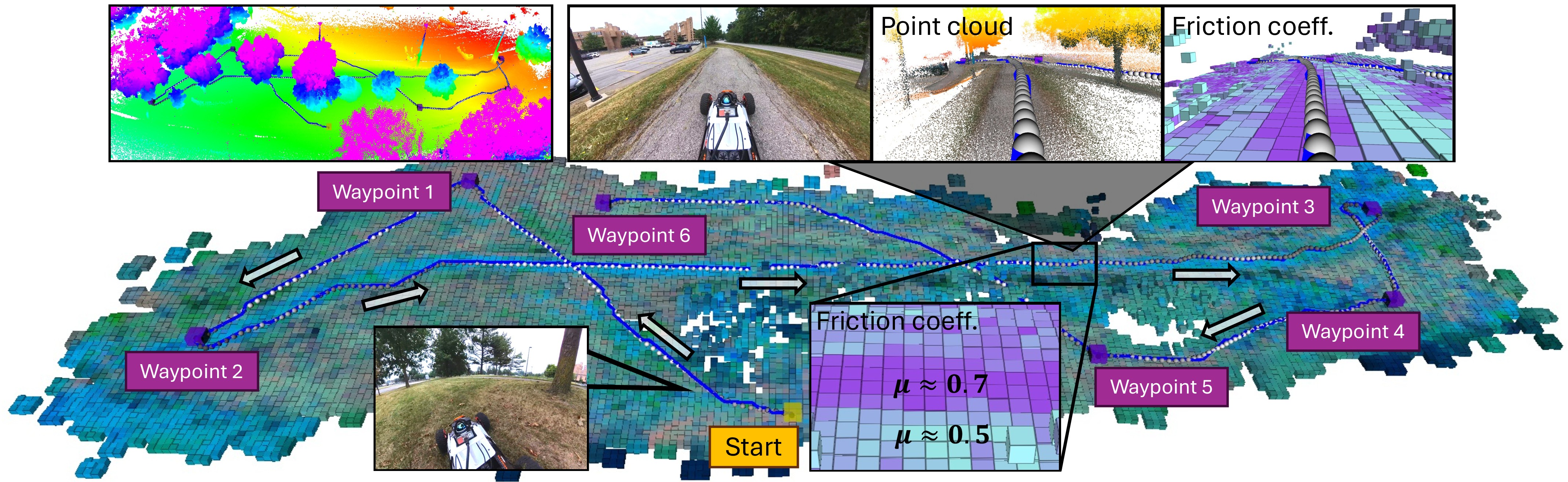}
    \caption{Example of navigating in a real-world curbside area from the start point to waypoints 1-6. The estimated friction coefficient is $\mu \approx 0.7$ on the asphalt road and $\mu \approx 0.5$ on the adjacent grass regions.}
    \label{fig:real_fm}
\end{figure*}

\subsubsection{Real-World Mapping and Navigation} \label{sec:realworld}

We validated the pipeline by scanning a lawn-and-road environment with a drone and deploying our navigation system on \textit{Rover}. \fref{fig:realworld}(a) shows the reconstructed point cloud generated using multi-view-structure (MVS) \citep{schoenberger2016mvs} from drone images. This colorized point cloud is then input into the mapping system for friction estimation as well as elevation and roughness computation. The resulting Stribeck layer is visualized in \fref{fig:realworld}(b), where the RGB channels correspond to Stribeck coefficients $\mu_v, \mu_d$ and $\mu_s$, respectively.
It is seen that the lawn and road can be clearly distinguished by their different friction properties. After the map is built, the planner deployed on \textit{Rover} generates trajectories to traverse the manually selected waypoints sequentially, as shown in \fref{fig:realworld}(c–e).

\subsection{Deployment in Unknown Environments}
\label{sec:nav_in_unknown}

Previous experiments validated the strong performance of AnyNav in known environments, which is critical for applications such as agricultural automation. However, in applications such as planetary exploration and disaster response, a prebuilt map is unavailable. In this section, we further evaluate AnyNav’s capability to incrementally construct a map using onboard sensors during exploration and to continuously replan based on the updated map.

\subsubsection{Quantitative Evaluation}
We conducted extensive testing of the online navigation in two mapping setups:

\myparagraph{Online Mapping from Scratch}
To rigorously test the reliability of AnyNav in unfamiliar environments, we adopt a mapping-from-scratch mode in which navigation starts with an empty terrain properties map at the beginning of each task. This setting presents a substantial challenge for the system: without any prior knowledge of the environment, the navigation system must simultaneously map the visible region during exploration, estimate friction and other properties in real time, and continually update its plans as new information becomes available. The limited range and resolution of perception sensors, along with obstacles that occlude portions of the view, further amplify the difficulty of reliable navigation.
For comparison, we select a representative model-based baseline using Model Predictive Path Integral (MPPI) \citep{williams2016aggressive} as the dynamics-aware planner. MPPI is widely used for planning in complex unstructured environments (e.g., \cite{sharma2023ramp, lee2023learning, cai2023probabilistic}), making it an appropriate baseline. For fairness, the MPPI planner receives the same terrain-property information as our method. It also incorporates friction limits in its dynamics model and applies roughness costs to avoid obstacles and highly uneven regions. In addition, we compare against a binary traversability baseline similar to \citep{ramirez2024real}, described in \sref{sec:quat_known_map}.
The experiments are conducted in the \textit{Garden} and \textit{Island} environments, using the same 50 tasks described in \sref{sec:quat_known_map}. To mitigate randomness, each task is executed three times to report average results. As shown in \tref{tab:mfs}, AnyNav achieves the highest success rate, surpassing the MPPI baseline by 16.3\% and the binary traversability baseline by 26.3\% on average.

\begin{figure*}[t]
    \centering
    \includegraphics[width=\linewidth]{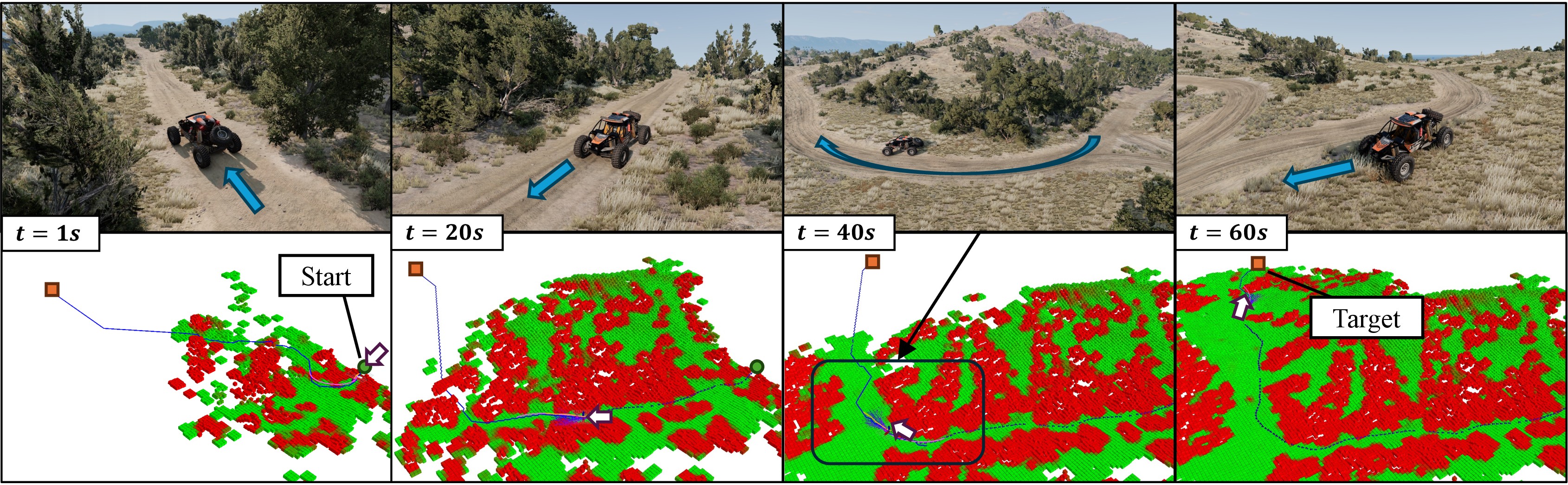}
    \caption{The roughness map is incrementally built from scratch during a navigation task in a dense vegetation region while the path is replanned accordingly. As the goal is on the right, the vehicle initially attempts to turn right at an early stage (e.g., $t=1s$). However, as it moves forward, the sensors continuously detect obstacles on the right, causing the planner to delay the turning point (e.g., $t=20s$), and the vehicle proceeds straight to avoid collisions. At $t=40s$, the planner identifies a flat region suitable for turning right. The vehicle is approaching the goal at $t=60s$ after the turn.}
    \label{fig:sim_rm}
\end{figure*}

\begin{figure*}[t]
    \centering
    \includegraphics[width=\linewidth]{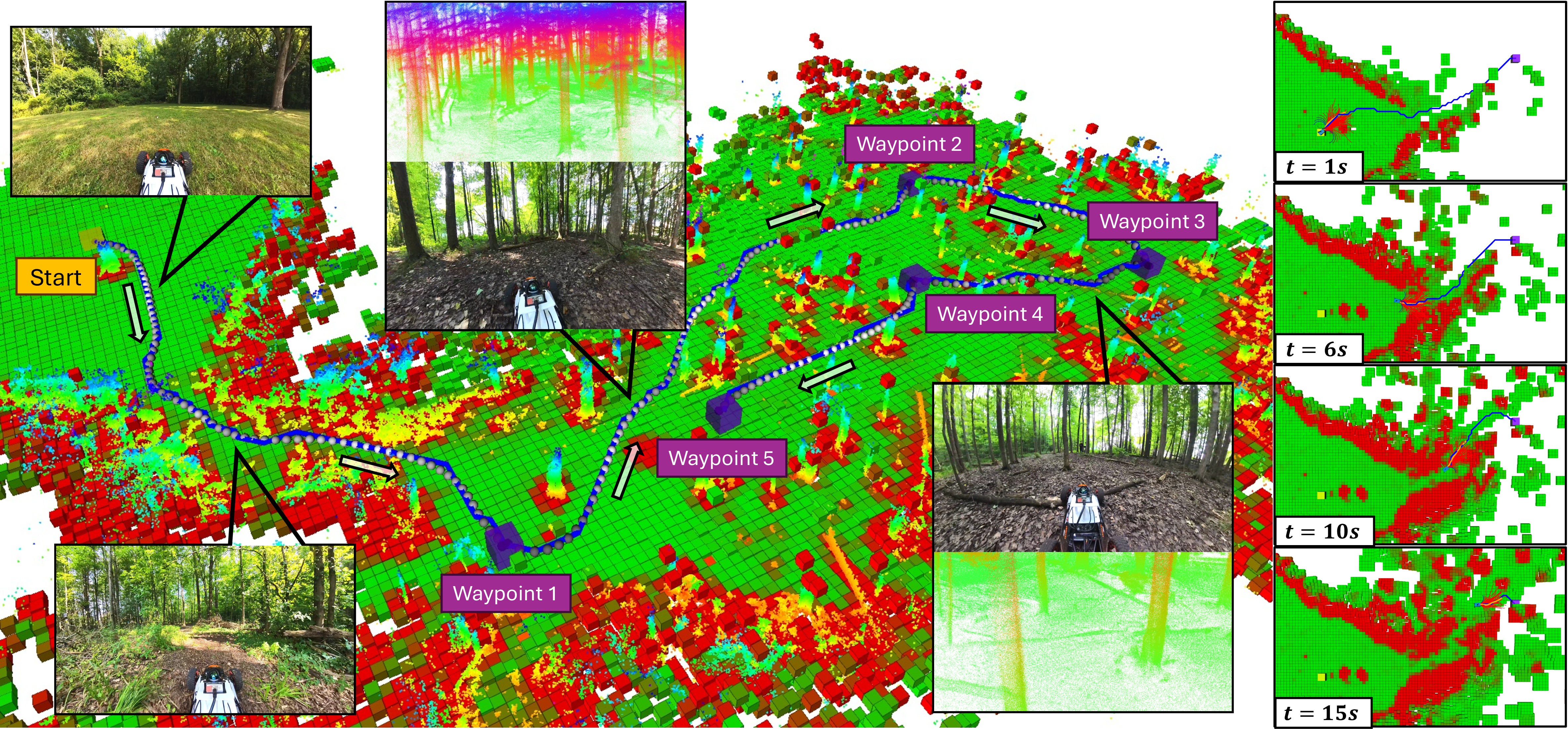}
    \caption{Real-world navigation in a forest environment with dense bushes and trees. The roughness map guides the vehicle to avoid collision. The right column shows the roughness map's building process as the vehicle drives toward waypoint 1.}
    \label{fig:real_rm}
\end{figure*}

\myparagraph{Online Accumulative Mapping}
A more practical setup is to retain the mapped regions, as they may benefit future tasks. In the accumulative-mapping mode, the map is progressively built across tasks, allowing later tasks to leverage the mapped terrain properties of areas explored earlier. Although the earliest tasks face challenges similar to those in the mapping-from-scratch mode, as the map becomes increasingly complete, later tasks can rely on more global information beyond the immediate visible range.
We use the same baselines, environments, and tasks as in the mapping-from-scratch evaluation. To reduce randomness, the task order is randomly shuffled three times, and each resulting sequence is executed once. The results are summarized in \tref{tab:am}. Compared with the MPPI baseline, AnyNav achieves 7.33\% and 32.67\% higher success rates in the two environments, respectively.

\begin{figure*}[t]
    \centering
    \includegraphics[width=\linewidth]{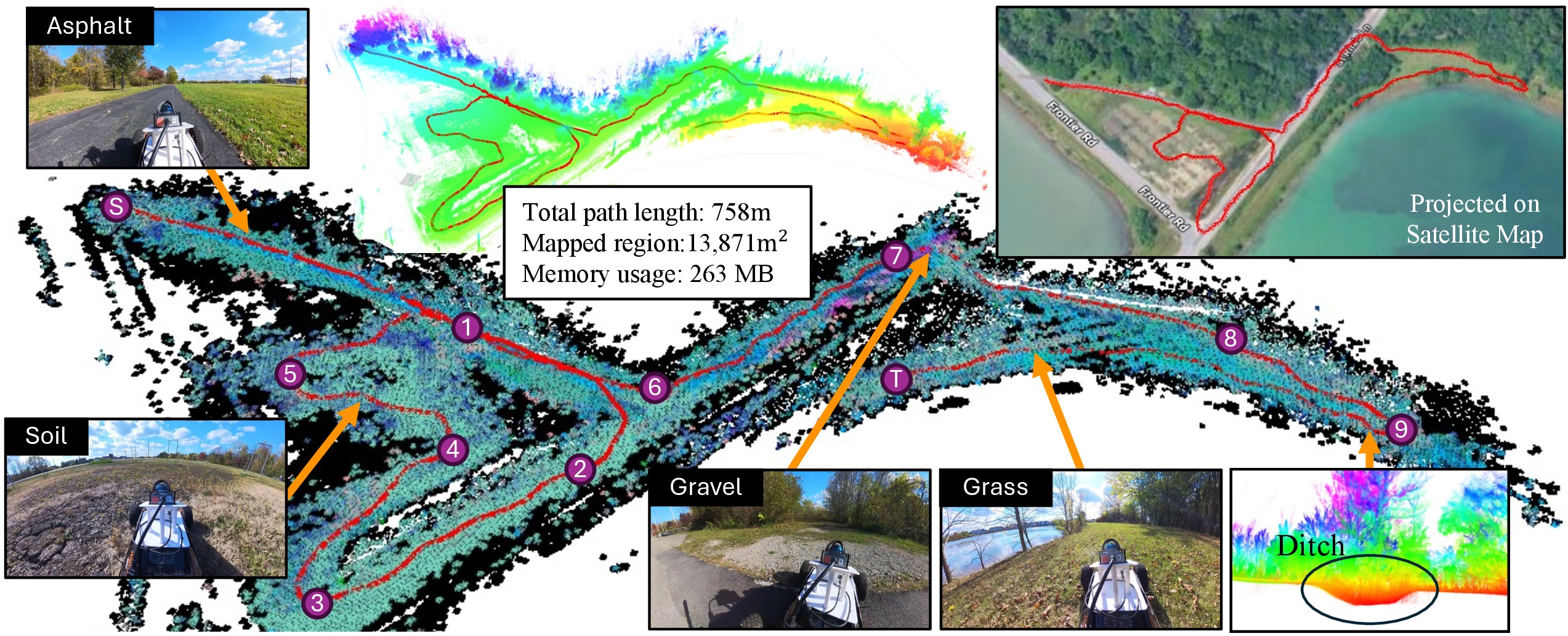}
    \caption{A long-distance navigation real-world test traversing diverse terrain types. The displayed Stribeck map is incrementally constructed during task execution. Purple waypoints are manually assigned, while red paths are generated by our method.}
    \label{fig:real_long2}
\end{figure*}

\myparagraph{Comparisons and Analysis}
Examining the performance of AnyNav across the two mapping setups in \tref{tab:mfs} and \tref{tab:am}, it is observed that the success rate is higher by 4\%-11\% in the accumulative mapping mode than in the mapping from scratch mode. This is because the gradually completed map assists later tasks in making better early-stage planning decisions, helping to avoid dead ends and hazardous conditions ahead. 
Besides, the success rate difference of AnyNav between the two modes is larger in the \textit{Garden} environment (11.33\% gap) than in \textit{Island} (4\% gap), since \textit{Garden} is specifically designed to include large slippery and inclined regions with only a single traversable path, which is difficult to find without prior mapping knowledge.

Compared with the baselines, we observe that in the \textit{Garden} environment, the MPPI baseline achieves a success rate close to AnyNav in the mapping-from-scratch mode (within 2\%), but it remains clearly worse than AnyNav in the accumulative-mapping mode (7.33\% gap). We attribute this to the low efficiency of action sampling in MPPI, which limits its effective planning horizon and prevents it from performing global planning that fully exploits far-range information in the progressively completed map. The binary traversability baseline consistently underperforms the other two methods, as it does not incorporate friction constraints, which are crucial in these challenging environments.

\subsubsection{Demonstration}
We next demonstrate representative navigation tasks in initially unknown environments, both in simulation and in real-world scenarios.

\myparagraph{Navigation in Friction-Challenging Environments} 
To reveal how real-time friction estimation fundamentally impacts navigation behavior, we demonstrate AnyNav in friction-challenging environments. 
\fref{fig:sim_fm} illustrates a task in which the goal position is on a hilltop, where the hill is covered by slippery icy regions and only one dirt road can provide enough friction for climbing. Four screenshots at different time steps clearly illustrate how the friction map is gradually built. It is observed that the path is continuously replanned and finally chooses the road after it is perceived and reasoned as a high-friction region. 
In the real world, \fref{fig:real_fm} presents tasks in which the vehicle traverses a sequence of manually selected waypoints in a curbside region. 
Notably, the Stribeck coefficients of the horizontal road in the middle differ clearly from those of the adjacent lawn. As a result, the task from waypoint 2 to 3 follows the road to benefit from its higher grip.

\myparagraph{Navigation in Obstacle-Dense Areas}
To demonstrate how terrain roughness mapping guides safe motion in cluttered environments, we deploy AnyNav in obstacle-dense areas.
\fref{fig:sim_rm} shows a longer navigation task in a dense vegetation area. We visualized the expansion of the roughness map over time. Initially, the vehicle attempted to turn right toward the goal. However, as it continuously detected dense vegetation on the right, the planner decided to proceed straight until it identified a flat region suitable for turning.
In the real world, \fref{fig:real_rm} shows navigation tasks in a complex forest environment. The construction process of the roughness map from the start to waypoint 1 is visualized on the right. The system enabled effective obstacle avoidance using the roughness map, allowing the vehicle to successfully navigate through the woods.

\myparagraph{Real-World Long-Distance Navigation}
To evaluate the robustness of AnyNav over extended missions, we conducted long-distance navigation tests in real-world environments.
\fref{fig:real_long2} illustrates a representative run covering 758 meters across 10 consecutive tasks. AnyNav successfully traverses diverse terrain types, including asphalt, mud, gravel, and grass, by selecting feasible routes, avoiding obstacles, and slowing appropriately at a ditch.
By the end of the mission, the terrain-properties map covered a 13,871m$^2$ region while occupying only 263 MB of memory, enabled by our instance-while-used mapping design. This experiment shows that AnyNav maintains reliable performance over long horizons, confirming its suitability for real-world applications.

\section{Limitation \& Conclusion}

In this paper, we present AnyNav, a self-supervised framework for friction learning and physics-informed off-road navigation.
We employ a neuro-symbolic approach to predict terrain friction coefficients from visual inputs, and achieve sim-to-real transfer learning in a self-supervised manner by grounding the neural model in symbolic physical reasoning through bilevel optimization.
The navigation system performs real-time terrain property mapping using onboard sensors and leverages the predicted friction knowledge for physics-informed path and speed planning. Experiments conducted in both simulation and real-world settings demonstrate the accuracy of our friction prediction model and the robustness of our navigation system.

Despite the demonstrated performance, several limitations remain to be addressed. First, although the predicted friction coefficients provide an indirect measure of terrain traversability and risk, their uncertainty distribution is not explicitly modeled. Incorporating probabilistic friction estimation into planning is an interesting direction for future work. Second, similar to human perception, which can make incorrect predictions from appearance alone, our visual friction estimation may also fail when terrains have nearly identical visual features but different mechanical properties. Future work could introduce complementary sensing modalities, such as thermal or infrared imaging, to provide information beyond visual appearance and help disambiguate such terrains. We also plan to extend the framework to predict additional terrain properties, such as stiffness, elasticity, and plasticity, and deploy it on full-scale all-terrain vehicles for autonomous off-road tasks.

\begin{acks}
This work was partially funded by DARPA award HR00112490426 and ONR award N00014-24-1-2003. The views and conclusions contained in this document are those of the authors and should not be interpreted as representing the official policies, either expressed or implied, of DARPA or ONR.
The authors also thank Bowen Li (Carnegie Mellon University) for the insightful discussions.
\end{acks}

\balance
\bibliographystyle{SageH}
\bibliography{citations}


\newpage
\appendix

\section{Weight Transfer Effect}
\label{ssec:weight-transfer}

\begin{figure}[b]
    \centering
    \includegraphics[width=0.8\linewidth]{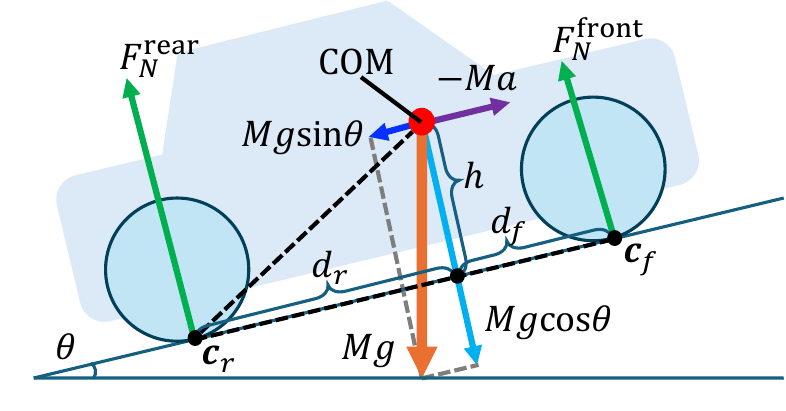}
    \caption{\textbf{Analysis of weight transfer of a vehicle on an inclined plane.} The vehicle accelerates down an incline at an angle \(\theta\). The gravitational force \(Mg\) acts on the COM, located at a height \(h\) above the slope. The normal forces at the front and rear wheels, \(F_N^{\text{front}}\) and \(F_N^{\text{rear}}\), act at contact points \(\mathbf{c}_f\) and \(\mathbf{c}_r\), respectively. The distances along the slope from the COM to these contact points are denoted as \(d_f\) and \(d_r\). The inertial force from the vehicle’s acceleration, \(-Ma\), and the components of gravitational force, \(Mg\sin\theta\) and \(Mg\cos\theta\), are also shown for analysis of weight transfer.}
    \label{fig:weight_transfer}
\end{figure}

When a vehicle contacts the ground, the ground exerts supporting normal forces on its wheels. The magnitude of the normal force on each wheel is influenced by the weight transfer effect \citep{su2023double}. The bicycle model~\citep{ailon2005controllability} is commonly used to quantify the distribution ratio of these forces along the longitudinal and lateral axes. For simplicity, we present the analysis along the longitudinal axis, noting that the same principles can be applied to the lateral direction.
As shown in \fref{fig:weight_transfer}, consider the vehicle accelerating at a rate $a$ on an upward slope inclined at an angle $\theta$. The gravitational force $Mg$ acts on the COM, which is elevated above the slope by a height $h$. 
The front and rear wheels' contact points are referred to as $\mathbf{c}_{f}$ and $\mathbf{c}_{r}$, corresponding to the exerted front normal force $F_N^{front}$ and rear normal force $F_N^{rear}$, respectively.
The distances along the slope from the COM to the front and rear contact points are denoted as $d_f$ and $d_r$, respectively. 
To analyze the system, we take $\mathbf{c}_r$ as the reference point and establish a 2D local frame at this contact point, where the two axes are parallel and perpendicular to the slope, respectively. In this non-inertial frame, an equivalent inertial force $-Ma$ acts on the COM due to acceleration. Applying the conservation of angular momentum w.r.t. $\mathbf{c}_r$ gives us
\begin{equation}
    (d_r+d_f)F_N^{front} + hMg\sin\theta + hMa - d_r Mg\cos\theta = 0,
\end{equation}
where the frictional forces are excluded, as they have zero moment arms relative to $\mathbf{c}_r$.
Solving for $F_N^{front}$, we obtain
\begin{equation}
    F_N^{front} = \frac{- hMg\sin\theta - hMa + d_r Mg\cos\theta}{d_r + d_f}.
\end{equation}
Similarly, $F_N^{rear}$ can be obtained by performing an analysis at $\mathbf{c}_f$, hence the ratio of the front-to-rear normal force is
\begin{equation}
    \beta \triangleq \frac{F_N^{front}}{F_N^{rear}} = \frac{- hMg\sin\theta - hMa + d_r Mg\cos\theta}{hMg\sin\theta + hMa + d_f Mg\cos\theta}.
\end{equation}
A similar process can be applied to compute the transverse force ratio $\gamma \triangleq F_N^{left}/F_N^{right}$. 

\section{Hyperparameters and weights}
\label{sec:hyperparam}

For better reproducibility, we provide the parameter values and weight coefficients used in both the training process and the planning system, along with insights into how the values of critical parameters were determined.

For training in simulation, we set $\lambda = 10$ in \eqref{eq:sim_tot_loss} to balance the contributions of the acceleration loss and the prior-knowledge loss, since $L_{accel}$ is observed to be approximately $10\times$ larger than $L_{prior}$ in our experiments. We use the Adam optimizer with \texttt{learning\_rate=5e-5}, \texttt{batch\_size=32}, and \texttt{weight\_decay=1e-4}. We train the model for approximately 250 epochs, until the loss and error indicators converge on the validation set.

For sim-to-real transfer, in the objective of lower-level optimization \eqref{eq:tot_il_cost}, we use $\mathbf{W}_a=\mathbf{I}_3, \mathbf{W}_\alpha=\mathbf{I}_3, \mathbf{W}_v=\mathbf{I}_3, \mathbf{W}_q=\mathbf{I}_3, \mathbf{W}_s=0.1\mathbf{I}_4,$ and $\mathbf{W}_w=0.01\mathbf{I}_3,$ where $\mathbf{I}_3$ and $\mathbf{I}_4$ denote $3\times3$ and $4\times4$ identity matrices, respectively. This configuration provides a good numerical balance among the cost terms.
We employ the Levenberg–Marquardt (LM) algorithm with a \texttt{Cholesky} solver and the \texttt{TrustRegion} strategy using a \texttt{radius=1e4}. Convergence is determined by a \texttt{StopOnPlateau} scheduler with \texttt{max\_steps=50}, \texttt{patience=2}, and \texttt{decreasing=1e-3}. Further implementation details can be found in~\cite{wang2023pypose}.

In the mapping system, we use a grid resolution of $L = 1.6\text{m}$ for full-size vehicles in simulation and $L = 0.4\text{m}$ for smaller real-world robots, which is slightly larger than the vehicle width. The overhanging object height $h$ is set to $2\text{m}$ in simulation and $0.5\text{m}$ in real-world experiments, slightly higher than the respective vehicle heights.

In the planning system, we use the weighted sum of the distance cost $C_d$, friction cost $C_f$, slope cost $C_\Delta$, roughness cost $C_r$, and steering cost $C_s$ to quantify the traversal effort:
\begin{equation} \label{eq:A_star_cost}
    C_{\text{PQ}} = \lambda_d C_d + \lambda_f C_f + \lambda_\Delta C_\Delta + \lambda_r C_r + \lambda_s C_s,
\end{equation}
where $\lambda_d=50, \lambda_f=1, \lambda_\Delta=10, \lambda_r=8, \lambda_s=1$. 
These values are manually tuned by observing the vehicle’s behavior in 10 debugging navigation tasks conducted in simulation and are kept fixed throughout all experiments. The debugging tasks are excluded from the reported experimental results to prevent bias in evaluation.

The parameters of the PD controller are shown in \tref{tab:pd}. These parameters were manually selected through small-scale testing to ensure that the vehicles can follow the planned trajectory smoothly. The \textit{Rover}'s motor includes reduction gears that provide strong damping; therefore, the brake and derivative (D) terms are not required.

\begin{table}[!htp]\centering
\caption{PD controller parameters.}\label{tab:pd}
\begin{tabular}{c|cc|cc}\toprule
&\multicolumn{2}{c|}{Simulation} &\multicolumn{2}{c}{Real-world} \\
&\textit{Racetruck} &\textit{Rockbouncer} &\textit{Rover} &\textit{Racer} \\\midrule
$P_\text{throttle}$ &0.5 &0.5 &0.5 &0.2 \\
$D_\text{throttle}$ &0.05 &0.05 &0 &0.02 \\
$P_\text{brake}$ &0.2 &0.5 &- &0.2 \\
$D_\text{brake}$ &0.02 &0.05 &- &0.02 \\
$P_\text{steering}$ &0.02 &0.05 &0.05 &0.02 \\
$D_\text{steering}$ &0.002 &0.005 &0 &0.002 \\
\bottomrule
\end{tabular}
\end{table}

\end{document}